\documentclass[sigconf,balance=false]{acmart}
\usepackage{popets}
\usepackage{footnote}
\makesavenoteenv{tabular}
\makesavenoteenv{table}
\setcopyright{popets}
\copyrightyear{YYYY}

\acmYear{}
\acmVolume{}
\acmNumber{}
\acmDOI{}
\acmISBN{}
\acmConference{}
\renewcommand\footnotetextcopyrightpermission[1]{} 
\usepackage{makecell}
\usepackage{subcaption}

\usepackage{bm}

\definecolor{oh_color_gray_dark}{HTML}{727272}
\definecolor{oh_color_gray_light}{HTML}{bdbdbd}
\definecolor{oh_color_green_leaf}{HTML}{22B14C}

\definecolor{oh_yellow}{HTML}{ffc022}
\definecolor{oh_green}{HTML}{009100}
\definecolor{oh_blue}{HTML}{005c94}
\definecolor{oh_purple}{HTML}{8421a9}
\definecolor{oh_red}{HTML}{dc0000}
\definecolor{oh_orange}{HTML}{ff6600}
\definecolor{oh_brown}{HTML}{804d00}
\definecolor{oh_black}{HTML}{0e232e}

\usepackage{tikz}
\newcommand{\tikzcircle}[2][red,fill=red]{\tikz[baseline=-0.5ex]\draw[#1,radius=#2] (0,0) circle ;} 

\usepackage{multirow}
\usepackage[normalem]{ulem}

\usepackage{enumitem}

\begin{document}
\title{Identifying Privacy Personas}

\author{Olena Hrynenko}
\orcid{}
\affiliation{%
  \institution{Idiap Research Institute}
  \city{}
  \country{}}
\affiliation{
  \institution{École Polytechnique Fédérale de Lausanne}
  \city{}
  \country{}
}

\email{olena.hrynenko@idiap.ch}

\author{Andrea Cavallaro}
\affiliation{%
  \institution{Idiap Research Institute}
  \city{}
  \country{}}
\affiliation{
  \institution{École Polytechnique Fédérale de Lausanne}
  \city{}
  \country{}
}

\email{a.cavallaro@idiap.ch}

\begin{abstract}
Privacy personas capture the differences in user segments with respect to one's knowledge, behavioural patterns, level of self-efficacy, and perception of the importance of privacy protection. Modelling these differences is essential for appropriately choosing personalised communication about privacy (e.g. to increase literacy) and for defining suitable choices for privacy enhancing technologies (PETs). While various privacy personas have been derived in the literature, they group together people who differ from each other in terms of important attributes such as perceived or desired level of control, and motivation to use PET.  To address this lack of granularity and comprehensiveness in describing personas, we propose eight personas that we derive by combining qualitative and quantitative analysis of the responses to an interactive educational questionnaire. We design an analysis pipeline that uses divisive hierarchical clustering and Boschloo's statistical test of homogeneity of proportions to ensure that the elicited clusters differ from each other based on a statistical measure. Additionally, we propose a new measure for calculating distances between questionnaire responses, that accounts for the type of the question (closed- vs open-ended) used to derive traits. We show that the proposed privacy personas statistically differ from each other. We statistically validate the proposed personas and also compare them with personas in the literature, showing that they provide a more granular and comprehensive understanding of user segments, which will allow to better assist users with their privacy needs. 

\end{abstract}


\maketitle

\section{Introduction}

People differ in their attitudes, perceptions, and expectations towards privacy ~\citep{zhao_privacyalert_2022, zerr_picalert_2012}. Being able to model these differences is a key step for providing personalised support~\citep{li_towards_2020}. Personas allow to model user segments~\citep{jansen_data-driven_2020} based on various attributes such as goals, needs~\citep{jansen_data-driven_2020}, preferences~\citep{colnago_is_2022}, concerns~\citep{milne_information_2017}, knowledge~\citep{ponnurangam__kumaraguru_privacy_2005, biselli_challenges_2022}, behaviours~\citep{dupree_privacy_2016, biselli_challenges_2022}, or financial wealth~\citep{stover_investigating_2023}. Personas should be as discriminative and realistic as possible~\citep{stover_how_2023}. Privacy personas should describe user segments that help design privacy and security tools~\citep{dupree_privacy_2016}. 

Common attributes defining privacy personas are level of concern about privacy (\citep{ponnurangam__kumaraguru_privacy_2005, elueze_privacy_2018}), level of knowledge (\citep{biselli_challenges_2022, dupree_privacy_2016}), one's behaviours (\citep{biselli_challenges_2022, dupree_privacy_2016, schomakers_typology_2019}) and attitudes (\citep{dupree_privacy_2016, schomakers_typology_2019}) towards privacy. While these attributes for privacy personas are commonly explored, there are a number of limitations. The attributes are often considered in isolation,  leading to a loss of information for privacy personas~\citep{biselli_challenges_2022, schomakers_typology_2019}. Examples include using only {\em level of concern} by Westin~\citep{ponnurangam__kumaraguru_privacy_2005}, or only behaviour and knowledge by \citet{biselli_challenges_2022}. 

To overcome these limitations, we propose a new way of modelling privacy personas from the responses to an interactive questionnaire~\citep{ferrarello_reframing_2022}. For modelling our privacy personas we use a mix of qualitative and quantitative approaches to facilitate a comprehensive understanding of the data~\citep{creswell_designing_2007}. Starting from questionnaire responses ~\citep{ferrarello_reframing_2022}, we perform coding, code manipulation~\citep{dupree_privacy_2016} and annotation to form participants' feature vectors. To account for differences in nature between answers to closed- and open-ended questions in a questionnaire, we propose a new dissimilarity measure for evaluating differences between the participants. Besides self-efficacy, PET's efficacy and willingness to use PET, we use a wide set of persona attributes (e.g., privacy protection importance perception, level of knowledge), which allow for a more comprehensive understanding of a persona. We propose a pipeline based on hierarchical clustering that ensures that the elicited personas are statistically different to each other (i.e., Boschloo's test of homogeneity of proportions ~\citep{boschloo_raised_1970}),  rather than ad-hoc approaches previously used in the literature~\citep{dupree_privacy_2016, elueze_privacy_2018, ponnurangam__kumaraguru_privacy_2005}. 

In summary, our main contributions are: 
\begin{itemize}
    \item The identification of eight privacy personas elicited with an interactive questionnaire~\citep{ferrarello_reframing_2022}, which incorporates responses about one's self-efficacy, willingness to use PET, alongside other privacy persona attributes, allowing for a comprehensive persona definition. 
    \item The proposal of a dissimilarity measure that takes the nature of closed and open-ended questions into account for calculating differences between the responses of participants.
    \item A pipeline for validation of the cluster splits in a hierarchical structure based on statistical significance tests~\citep{boschloo_raised_1970}.
   \item The comparison of the identified personas with the privacy personas by Westin~\citep{ponnurangam__kumaraguru_privacy_2005}, 
   \citet{biselli_challenges_2022} and Dupree et al.~\citep{dupree_privacy_2016}, showing that our personas have a higher level of granularity, resulting in a more accurate representation of the population, which translates into better-aligned support for personas.  
\end{itemize}

\begin{table*}[t!]
\renewcommand{\arraystretch}{1.5}
\small
\begin{tabular}{p{0.03\linewidth}p{0.03\linewidth}p{0.37\linewidth}p{0.07\linewidth}p{0.09\linewidth}p{0.07\linewidth}p{0.07\linewidth}p{0.09\linewidth}}
\hline 

\multicolumn{1}{c}{} & \multicolumn{1}{c}{} & \multicolumn{1}{c}{} & \multicolumn{2}{c}{\textbf{\vspace{3px}Generation set}} & \multicolumn{3}{c}{\textbf{\vspace{-7px}Validation set}} \vspace{5px} \\ \cmidrule(lr){4-5} \cmidrule(lr){6-8}

\multicolumn{1}{c}{\multirow{-2}{*}{\textbf{\vspace{2px}\rotatebox{0}{Ref.}}}} & \multicolumn{1}{c}{\multirow{-2}{*}{\textbf{\vspace{2px}Method}}} & \multicolumn{1}{c}{\multirow{-2}{*}{\textbf{\vspace{2px}Privacy personas}}} & \multicolumn{1}{c}{\textbf{size}} & \multicolumn{1}{c}{\textbf{demogr.}} & \multicolumn{1}{c}{{\textbf{SD}}} & \multicolumn{1}{c}{\textbf{size}} & \multicolumn{1}{c}{\textbf{demogr.}} \\ \hline

\makecell[c]{\citep{ponnurangam__kumaraguru_privacy_2005}} & \makecell[c]{QN} & Fundamentalist, Pragmatist, Unconcerned & \multicolumn{1}{c}{N/A} &\multicolumn{1}{c}{N/A} & \multicolumn{1}{c}{N/A} & \multicolumn{1}{c}{N/A} & \makecell{N/A} \\ \hline

 \makecell[c]{\multirow{2}{*}{{{\citep{elueze_privacy_2018}}}}}&  \makecell[c]{\multirow{2}{*}{{{QL}}}}  & Fundamentalist, Intense Pragmatist, Relaxed Pragmatist, Marginally Concerned, Cynical Expert & \multicolumn{1}{c}{\makecell[c]{\multirow{2}{*}{40}}} &  \multicolumn{1}{c}{\makecell[c]{\multirow{2}{*}{\makecell{65-91 yo \\ British Canadians}}}} & \multicolumn{1}{c}{\makecell[c]{\multirow{2}{*}{N/A}}} & \multicolumn{1}{c}{\makecell[c]{\multirow{2}{*}{N/A}}} & \multicolumn{1}{c}{\makecell[c]{\multirow{2}{*}{N/A}}} \\ \hline

\makecell[c]{\citep{biselli_challenges_2022}} & \makecell[c]{QN} & Fundamentalist, Pragmatist, Unconcerned & \multicolumn{1}{c}{332} & \multicolumn{1}{c}{\makecell{18-75 yo \\ German}} & \multicolumn{1}{c}{no} & \multicolumn{1}{c}{324} & \makecell{18-75 yo \\ German} \\ \hline

\makecell[c]{\multirow{2}{*}{\citep{dupree_privacy_2016}}} & \makecell[c]{\multirow{2}{*}{HY}} & Fundamentalist, Technician, Amateur, Marginally Concerned, Lazy Expert & \multicolumn{1}{c}{\makecell[c]{\multirow{2}{*}{32}}} & \multicolumn{0}{c}{\makecell[c]{\multirow{2}{*}{\makecell{22-35 yo\\ North American}}}} & \multicolumn{1}{c}{\makecell[c]{\multirow{2}{*}{yes}}} & \multicolumn{1}{c}{\makecell[c]{\multirow{2}{*}{200}}} & \multicolumn{0}{c}{\makecell[c]{\multirow{2}{*}{\makecell{18-65 yo \\ American}}}}\\ \hline

\makecell[c]{\citep{schomakers_typology_2019}} & \makecell[c]{QN} & Guardian, Pragmatist, Cynic  & \multicolumn{1}{c}{337} & \multicolumn{1}{c}{\makecell{above 14 yo \\ German}} & \multicolumn{1}{c}{N/A} & \multicolumn{1}{c}{N/A} & \makecell{N/A} \\ \hline

\makecell[c]{\multirow{1}{*}{Ours}} & \makecell[c]{\multirow{3}{*}{HY}} & Knowledgeable Optimist, In-control Adopter, In-control Sceptic, Knowledgeable Pessimist, Helpless Protector, Occasional Protector, Adopting Protector, Unconcerned & \multicolumn{1}{c}{\makecell[c]{\multirow{3}{*}{130}}} & \multicolumn{1}{c}{{\makecell[c]{\multirow{3}{*}{\makecell{25-35, \\ British}}}}} & \multicolumn{1}{c}{\makecell[c]{\multirow{3}{*}{yes}}} & \multicolumn{1}{c}{\makecell[c]{\multirow{3}{*}{50}}} & {\makecell[c]{\multirow{3}{*}{\makecell{25-35, \\ British}}}} \\ \hline

\end{tabular}
\caption{Related work, methods that were used for eliciting privacy personas, and privacy personas. Key -- QN: quantitative, QL:~qualitative, HY: hybrid, SD: separate dataset, yo: years old.}
\label{tab:related_work_personas}
\end{table*}

The paper is organised as follows. Section~\ref{sec:related_work} covers previous methods used to define privacy personas. Section~\ref{sec:dataset_ferrarello} describes the dataset we used for personas elicitation. Section~\ref{sec:privacy_traits} presents how we form personas' traits with open coding. Section~\ref{sec:dissimilarity_measure} explains the new dissimilarity measure and how we elicit personas.
Section~\ref{sec:privacy_personas} describes our personas, which are then validated in Section~\ref{sec:validation}. Section \ref{sec:limitations} discusses the limitations of our approach, whereas
Section~\ref{sec:other_privacy_personas} maps our personas with the personas in the literature. Finally, in Section~\ref{sec:conclusion} we draw conclusions and describe future work.  

\section{Related Work}
\label{sec:related_work}

In this section, we discuss qualitative, quantitative, and hybrid methods for identifying privacy personas.

{\em Qualitative} methods generally extract codes from interview transcripts, and group the codes into themes that represent the main ideas from the data~\citep{braun_thematic_2022, corbin_grounded_1990}. The goodness measure is \textit{saturation}, namely "the point in data collection when no additional issues or insights emerge from data and all relevant conceptual categories have been identified, explored, and exhausted"~\citep{hennink_code_2017}. Qualitative methods allow to discover new concepts and build the basis of the persona features, accommodating the emerging concepts. 
{\em Quantitative} methods follow a pipeline that includes question construction, data collection, feature vector construction, and personas elicitation. \textit{Questions construction} generates a list of questions to ask the participants. This step is based on the findings of qualitative analysis. The list of questions is refined based on feedback from experts or focus groups (e.g., removing unclear or leading questions), or on statistical methods conducted on preliminary data. Note that the outcome of the questions construction step is a fixed set of closed-ended questions. The \textit{data collection} step consists of reaching out to participants and collecting data. \textit{Feature vector construction} translates the collected responses into the numerical feature vectors. \textit{Personas elicitation} involves using quantitative methods for grouping participants into personas. 
With {\em hybrid} methods, the outcome of \textit{questions construction} is a mixture of closed- and open-ended questions, which could later on be enriched with additional questions (e.g., in the case of semi-structured interviews).  Including open-ended questions means that to convert a participant's response into a feature vector, further qualitative analysis is required (e.g.,~code extraction).  Methods and corresponding privacy personas are summarised in Tab. \ref{tab:related_work_personas} and described below. 

Privacy personas by Westin~\citep{ponnurangam__kumaraguru_privacy_2005} are often used as a baseline for the privacy personas, despite being criticised for the lack of predictive power for the behaviours and knowledge~\citep{king_taken_2014}. Westin~\citep{ponnurangam__kumaraguru_privacy_2005} conducted more than 120 studies to explore the privacy concerns of people~\citep{us_government_publishing_office_opinion_2001} and identified three privacy personas~\citep{ponnurangam__kumaraguru_privacy_2005}, namely the {Fundamentalist}, the Unconcerned, and the Pragmatist. The {Fundamentalist}
feels very strongly about privacy matters, and supports new laws that allow for privacy regulations. This persona does not trust companies that ask for personal information, and chooses privacy protection over potential benefits. The Fundamentalist is worried about the use of their information, pessimistic about the future of privacy protection, and feel they have lost a lot of their privacy. The {Unconcerned}
does not feel that their privacy is violated. This persona feels less anxious about how others use information about them. The Unconcerned trusts organisations collecting personal information about them, and the benefits they may receive by revealing their information carry higher weight than the potential privacy harm they might face. The {Pragmatist} feels strongly about privacy and seeks fair information practices, weighing the benefits against the level of intrusiveness. This persona wants to have the freedom of choice to opt-out. For the Pragmatist, business organisations should earn the trust, rather than have it unconditionally. We will compare our personas with Westin's personas in Section \ref{subsec:westin_vs_ours}.

\citet{elueze_privacy_2018} investigated the differences in attitudes and concerns about online privacy in the lives of older adults. They conducted in-person interviews, coded the data and used the categorisation by Westin~\citep{ponnurangam__kumaraguru_privacy_2005} as a basis for their personas. Based on the key responses to privacy questions, people were assigned to the Fundamentalist, the Pragmatist or the Unconcerned classes. Then based on further analysis of the codes and through analysis of negative cases, the Pragmatist persona was divided into two groups, namely the Intense Pragmatist and the Relaxed Pragmatist. An additional group, the Cynical Expert, was established. 

\citet{biselli_challenges_2022}  identified privacy personas using a set of closed-ended questions that measure the knowledge and behaviour of participants. They showed a strong positive correlation between knowledge and privacy behaviour: the more people are aware of the threat, the more cautious they are. \citet{biselli_challenges_2022}  segmented people into three personas, adopting the naming convention of Westin~\citep{ponnurangam__kumaraguru_privacy_2005}. However, as the authors acknowledged, no formal validation on linking their personas to Westin's has been done. We will compare our personas with Biselli's in Section \ref{subsec:biselli_vs_ours}.

\citet{dupree_privacy_2016} measured people's attitudes and behaviour towards security practices. They used closed-ended questions for a questionnaire and open-ended questions for the semi-structured interviews for their \textit{data collection} step. Because \citet{dupree_privacy_2016} used the hybrid method, their \textit{feature vector construction} step involved manual coding and annotation.  The personas were generated by clustering participants' most discriminative \textit{traits} (participants' features): a \textit{trait} shared by multiple clusters was removed, and the participants were re-clustered.
Once the clusters were established, the unique cluster \textit{traits} were used to describe the personas.
To position personas in a motivation-knowledge space, they manually annotated the participants' responses with respect to their knowledge (low, medium, high), and motivation\footnote{Motivation is defined as: "the effort they [people] expend to protect their privacy or security"~\citep{dupree_privacy_2016}.} (low, medium, high) after the clusters were formed. \citet{dupree_privacy_2016} then calculated the overall knowledge and motivation per cluster. 
Dupree's Fundamentalist has high motivation and high knowledge, the Marginally Concerned has low motivation and low knowledge, the Lazy Expert has high knowledge and low motivation, the Struggling Amateur has medium motivation and medium knowledge, and the Technician has high motivation and medium knowledge.  

While the traits per cluster were considered unique, there is no information on how common they were within the cluster, e.g., how many participants shared a similar opinion. Hence, we claim that Dupree's knowledge-motivation persona assignment has a higher validity than the clusters' description which comes from the unique traits only. Differently to \citet{dupree_privacy_2016}, we use an idea of statistically significant differences between the clusters and introduce a new dissimilarity measure which treats closed- and open-ended questions differently. We will compare our personas with Dupree's personas in Section \ref{subsec:dupree_vs_ours}.

\citet{schomakers_typology_2019} conducted the interviews to select a set of questions for their follow-up quantitative study for privacy personas generation. They clustered participants based on their level of concern and protective behaviours\footnote{The privacy behaviour (PB) and privacy concern (PC) questions were~\cite{schomakers_typology_2019}: {\em I use every option that I know to protect my online privacy (e.g., deleting cookies, anti-virus software)} (PB1);  
{\em I specifically search for more options to protect my online privacy} (PB2); {\em  I use the default settings of my devices and applications without changing them. (rev)} (PB3); {\em I use the default settings of my devices and applications without installing additional software to protect my privacy. (rev) } (PB4); {\em In general, I am concerned about my privacy when I am using the internet. } (PC1); 
    {\em With some types of information collected on the internet I do not feel comfortable} (PC2); and 
    {\em I do not see risks when providing data on the internet. (rev)} (PC3).}, and identified the Privacy Guardian (the highest level of concern, the highest level of proactive behaviour), the Privacy Cynic (the lowest level of behaviour, moderately high level of concern), and the Privacy Pragmatist (lowest level of concern, moderately high level of protective behaviour).

A key difference with our method is that we use a tightly intertwined mixture of qualitative and quantitative analysis, not separating one from another, whereas \citet{schomakers_typology_2019} use the findings of the interviews for developing a closed-ended questionnaire. Since privacy personas trends could change over time~\citep{elueze_privacy_2018}, our personas are more suitable for allowing to model these changes. 

Moreover, we consider the relationship not only between the behaviour and concern attributes but also between a wider set of attributes, which allows for a more granular persona understanding. We will compare our personas with personas by Schomakers et al.~\citep{schomakers_typology_2019} in Section \ref{subsec:schomakers_vs_ours}.

\section{Questionnaire}
\label{sec:dataset_ferrarello}

We use a dataset provided by \citet{ferrarello_reframing_2022} with privacy stimuli and an educational element. This dataset measured one’s change in predisposition towards privacy once they learned more from the interaction. An answer to a question may be a sentence (e.g., "Privacy is the ability to share things online yet keep unwanted attention away.") or a phrase (e.g., "Moderately easy"). 

\subsection{Questions and Domain of Answers}

The questionnaire is composed of questions $q_j$, where $j = 1, \ ..., \ $ $Q~=~19$. 
Through the questionnaire, participants learnt about the existence of inferences on the images they share online. 
They were firstly asked a set of questions about privacy protection importance, their preferences, expectations about privacy, and their awareness about inferences on their images ($q_1$ -- $q_{10}$). Then participants were asked to upload the most recent image that they have shared online with their audience. Participants were then informed of the existence of a different party (i.e., a company) having access to data inferred from the images. They were shown a demo of how much information can be extracted from the image. We refer to this process as the first privacy stimulus. After the first privacy stimulus participants were asked how this information extraction makes them feel, if they understand what is happening to their data, and their perceived self-efficacy ($q_{11}$ -- $q_{14})$. Then the participants interacted with an image filter that acted as a Privacy Enhancing Technology (PET), a second privacy stimulus. Finally, participants' willingness to use this PET and perceived efficacy of the PET were measured, and the updated view on privacy protection was recorded ($q_{15}$ -- $q_{19}$).
The educational, interactive element allows to align participants with respect to an otherwise vague concept of privacy~\citep{gallardo_speculative_2023, schomakers_typology_2019} as well as to measure participants' self-efficacy, perception of the response efficacy, and motivation to use PET.

The questions ($q_{*}$) and the domain of answers to closed-ended questions ($a_{*}$) are provided below.

\begin{enumerate}[label = $q_{\arabic*}$] \small
    \item First of all, how important do you think it is to protect private information when sharing images online?
    \begin{enumerate}
    \item[$a_1$] very important | important | neutral | not that important | unimportant
    \end{enumerate}

    \item Image 1: cat (What personal information is it revealing?)
    \item Image 2: van (What personal information is it revealing?)
    \item Image 3: rollercoaster (What personal
information is it revealing?)
    \item What could you do to protect the personal information
in these images?
    \item Now look at the last 10 images you shared online. What
pieces of personal information do you find in those
images?
    \item How easy do you think it is to protect the personal
information you identified with what is currently available to you?
\begin{enumerate}
     \item[$a_7$] extremely easy | mostly easy | moderately easy | slightly easy | not at all easy
     \end{enumerate}
    \item How often would you like to be able to have control over
the personal information that you have identified?
\begin{enumerate}
    \item[$a_8$] always | often | sometimes | rarely | never
    \end{enumerate}
    \item Did you know that social media platforms check the images you share for sensitive content?
    \begin{enumerate}
    \item[$a_9$] yes, familiar with how this is done | yes, but don't understand how | no
    \end{enumerate}
    \item But they may not always stop there. Did you know that,
behind every picture that you share, pieces of personal
information can be found and used to generate a
personal profile that is used to control what you see?
\begin{enumerate}
    \item[$a_{10}$] yes, familiar with how this is done | yes, but don't understand how | no
    \end{enumerate}
    \item[]\textit{Privacy stimulus 1}
    \item How does this make you feel?
    \item Do you feel like you understand what happens to your personal information when you share an image online?
    \begin{enumerate}
    \item[$a_{12}$] fully | mostly | moderately | slightly | don't understand
    \end{enumerate}
    \item What could you do to protect the personal information
you selected?
    \item Do you think you have control of your personal information when sharing images online?
    \begin{enumerate}
    \item[$a_{14}$] can fully | mostly | moderately | only little | cannot control
    \end{enumerate}
    \item[] \textit{Privacy stimulus 2}
    \item Do you feel like the filter would allow you to protect your privacy?
    \begin{enumerate}
    \item[$a_{15}$] yes | unsure | no
    \end{enumerate}
    \item If these new filters were available, would you use them?
    \begin{enumerate}
      \item[$a_{16}$] would always | often | sometimes | rarely | never use these filters
      \end{enumerate}
    \item Why?
    \item Think back on everything you learnt about sharing your images online. How important do you think it is to protect the privacy of the pictures you share online?
        \begin{enumerate}
        \item[$a_{18}$] very important | important | neutral | not that important | unimportant
        \end{enumerate}
    \item Finally, how would you now define the term privacy
when related to online images?
\end{enumerate}

\begin{figure*}[t]
    \centering
    \includegraphics[width = 1\linewidth]{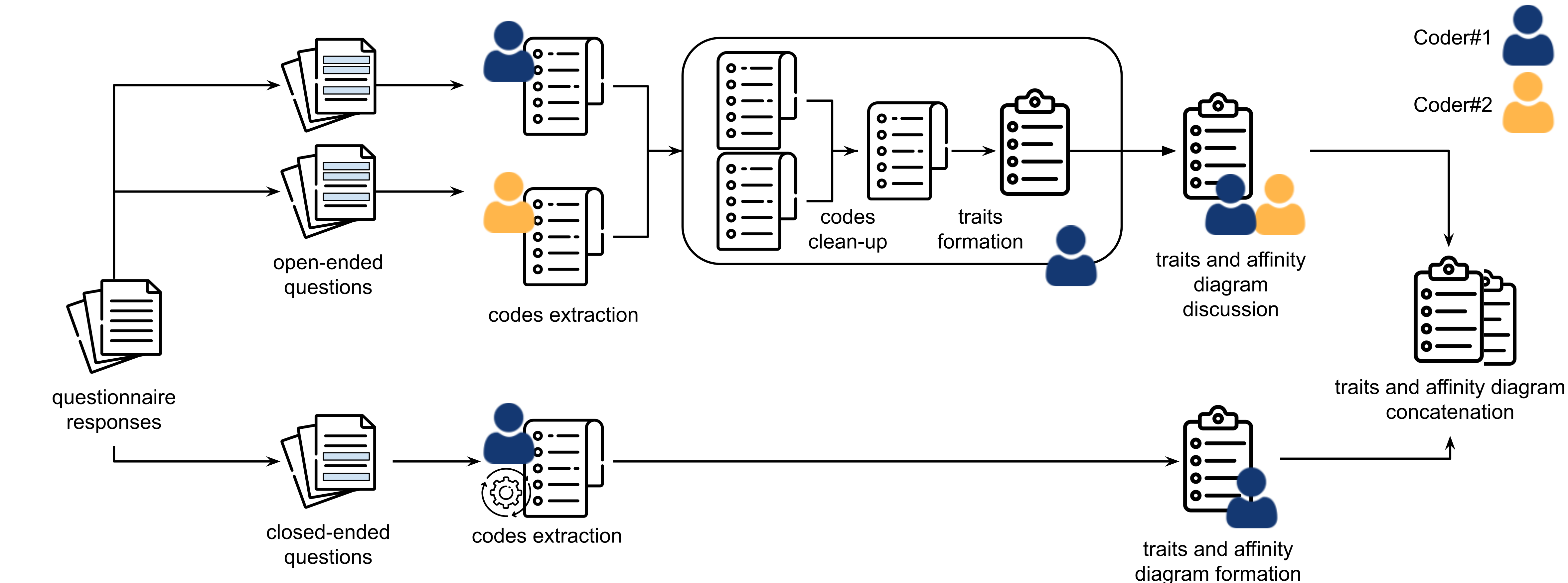}
    \caption{Process of extracting codes and traits formation. We follow different processes for open and closed-ended questions. Coding and traits extraction for open-ended questions consists of code extraction (done by two coders independently); code clean-up, traits formation (done by one coder); traits and affinity diagram discussion (done by two coders). For closed-ended questions the Likert-scale answers were the codes themselves (however coder paraphrased them to full sentences), traits and affinity diagram formation. An affinity diagram helps to organise data which initially seems unstructured~\citep{lucero_using_2015}, allowing to find categories of the traits~\citep{dupree_privacy_2016}. At the end of the process, there is a concatenation of the affinity diagram and the extracted traits.}
    \label{fig:from_quest_to_traits}
\end{figure*}

\subsection{Responses}
The dataset by \citet{ferrarello_reframing_2022} consists of 200 responses to an interactive questionnaire given by participants from the UK and was collected using  Prolific~\citep{prolific_prolific_2024}. The recruited participants were users of one of the following social media: Facebook, Twitter, Instagram, and Snapchat. The participants had at most 20 submissions on Prolific, and had their income in the range of £10,000-£50,000. The dataset is gender balanced, and the age of the participants ranged from 25 to 35 (mean: 30.2, standard deviation: 2.8). Participants were paid 7.10~GBP for taking part in a questionnaire. The study by \citet{ferrarello_reframing_2022} was approved by the ethics board. 
 
 We represent a participant $i$, $P_i$, through their answers to the questionnaire. We used all questionnaire responses available, except four, because they had missing answers. We used 64 responses for coding and 180 for annotation. For generating personas we used $G = 130$ responses that were not previously used for coding. We call these responses the generation set, $\mathscr{G}$. For the validation set, $\mathscr{V}$, we re-used $V = 50$ questionnaire responses that were previously used for coding.

\subsection{Biases and Priming}  The order of the questions of the questionnaire~\citep{ferrarello_reframing_2022} is important. Questions asked after the first {\em privacy stimulus} aimed at measuring the effect of being exposed to a personal information inference demo. Similarly, questions after the second {\em privacy stimulus} aimed at measuring the willingness of a participant to use the introduced PET. Finally, there were questions that measured the same attribute (i.e., privacy protection importance) at the beginning and at the end of the questionnaire to evaluate the effect on a participant of the educational interactive element (i.e.~being introduced to privacy-related threats, being offered the PET to protect personal information). 

The existence of an interactive element also introduced the bias of the personalised element. One of the biases of the questionnaire is that we only observe the self-reported behaviours and future decisions, which might not correspond to the actual behaviour. For example, the decision to often use filter ($q_{16}$) might be different to the actual observed behavioural patterns~\citep{biselli_challenges_2022}.

\begin{table*}[t!]
\renewcommand{\arraystretch}{1.5}
\footnotesize
\centering
\begin{tabular}{p{0.3\textwidth}p{0.6\textwidth}}
\hline
codes from open-ended questions  & codes from closed-ended questions  \\ \hline
would use filters to avoid relevant info being sold                  & initially expressed a desire to always be in control of their personal information                                                                                                       \\
I would use filters to avoid being objectified                       & initially does not have an opinion on privacy protection                                                                                                                                 \\
would use filters not to share things with strangers                & illustrated no awareness of social media being to extract personal information from an image and generate a personal profile on a user                                                   \\
filters "throw off any algorithm"                                    & after being exposed to how companies can extract data from their images they feel that they have moderate control over personal information when sharing images online                   \\
filters deteriorate colours & significant mismatch between the perceived and desired level of control over personal information after being exposed to companies being able to
extract personal information \\
filters protect information about me                                 & they would sometimes use the proposed privacy enhancing filter when sharing images online                                                                                              \\
\hline                      
\end{tabular}
\caption{Examples of the codes extracted from closed-ended and open-ended questions.}
\label{tab:codes}
\end{table*}

\section{Feature Vector Construction}
\label{sec:privacy_traits}

The dataset includes information from closed-ended and open-ended questions. For a feature vector construction, we use open coding through Grounded Theory~\citep{gibbs_analyzing_2018} to extract codes, which are sentences or phrases that propagate one idea~\citep{braun_thematic_2022}. We then group codes into traits, which are low-level groupings of codes that are summarised with a phrase or a sentence~\citep{dupree_privacy_2016}. To prepare for annotation, we form an affinity diagram~\citep{lucero_using_2015} of traits. Finally, we perform annotation. Fig. \ref{fig:from_quest_to_traits} shows the process of code extraction, and trait and affinity diagram formation.

\subsection{Code Extraction}

 We randomly sample 64 questionnaire responses for code extraction from the open-ended questions. This is twice the number of participants on which \citet{dupree_privacy_2016} have identified their personas.  We adopt a random sampling approach to avoid any human bias in the sampling process.  

\noindent \textbf{Codes from open-ended questions.} The process of extracting codes for open-ended questions consists of independent code extraction followed by code cleaning. 

Coders $C1$ and $C2$ extract the codes from the questionnaire responses without consulting each other to avoid introducing bias at this stage. The process involves reading the questionnaire responses and extracting atomic pieces of meaning -- phrases that cannot be subdivided into smaller ideas. Here we follow advice on extracting codes by answering the W's and H's questions~\citep{gibbs_analyzing_2018}: who, why, when, what and how, how much, and how long? For example, one of the answers to $q_{11}$ was "\textit{Very uneasy, not entirely surprised. Confirmed why I don't post often on social media}". This answer generated multiple codes: "feels uncomfortable", "feels not surprised", and "claims does not post often".  

We then create a list of codes, where each code is unique. The codes provided by each coder were concatenated and further processed. $C1$ removed duplicates of the codes which were caused by slight variability in annotation. For example, codes "high level of knowledge" and "highly knowledgeable" were mapped to the same code. Additionally, due to inevitable human error in the initial coding, the codes that encapsulated two or more distinct ideas were separated into multiple codes. For example, the code "filters serve a dual purpose: nice image + protection" was split into "filters increase the quality of the image" and "filters allow to protect me". 

\noindent \textbf{Codes from closed-ended questions.}
Each answer to closed-ended questions is a code once it is paraphrased. We paraphrase the answer for it to be understood without reading the corresponding question ("I am neutral" answer to $q_1$ is paraphrased to "initially does not have an opinion on privacy protection"). More examples of the closed-ended codes are in Tab.~\ref{tab:codes}. Coding and paraphrasing of closed-ended responses enable Boschloo's test~\citep{boschloo_raised_1970} and ensure that codes are understood without additional context. 

Additionally, we generate codes that capture the change of privacy protection importance ($q_1$ and $q_{18}$), and the desired and perceived level of control mismatches ($q_8$ and $q_{14}$) to identify the mismatch between perceived and desired levels of control. 

\subsection{Traits}
Following the approach proposed by \citet{dupree_privacy_2016}, after code extraction, we form traits and build an affinity diagram~\citep{lucero_using_2015}. We form traits to ensure that when we describe participants, we refer to an idea (i.e., \textit{trait}) rather than an instance of an idea (i.e., \textit{code}).
For example, in $q_{17}$ one of the participants expressed that they would not use the proposed PET. They explained their decision by saying that the "keywords were not erased". This answer generated a code  "expressed an opinion that keywords can be erased". Here the participant was not aware that a machine learning pipeline that is trained to extract concepts from an image would always produce an output, it is the quality of the output that could be altered. So code "expressed an opinion that keywords can be erased" was mapped to a trait "lacking knowledge in some aspects". Other codes that were mapped to the same trait are "thought that it is possible to change the automatically generated keywords by hand", and "expressed an opinion that setting profile to private would help with preventing the inference on their images". 

For the codes elicited from the open-ended codes,  $C1$ manually performed low-level clustering on codes, forming the traits. Then, the traits were discussed and agreed on by $C1$ and $C2$, following the approach by \citet{dupree_privacy_2016}. As a result, 75 traits were generated from answers to open-ended questions. Since closed-ended codes are different to each other by design (i.e., different response anchors of the Likert scale), we adopt a one-to-one mapping between the codes and the traits. We derive 58 traits from answers to closed-ended questions.  A set of all traits $\mathcal{T}$ consisted of $T = 133$ traits. 

After the traits were formed, $C1$ grouped them with respect to higher-level categories~\citep{dupree_privacy_2016} by creating an affinity diagram. Forming an affinity diagram allows us to mimic the axial coding step in Grounded theory. The aim of the affinity diagram is to reduce the cognitive load for annotators later in the process. Identified higher-level categories include the participant's level of knowledge, the emotions after the first stimulus, and categories of personal or private data.  Finally, $C1$ and $C2$ discussed the coherence of both the elicited traits, and the affinity diagram, and resolved disagreements.  

\subsection{Annotation}
By considering the traits in $\mathcal{T}$, we form a basis of the feature space for participants. Each participant $P_i$ can be represented in the form of $\bm{p_i} \in \{0, 1\}^T$. The $n$'th value of $\bm{p_i}$ is equal to one if $P_i$ propagated trait $t_n$ in their questionnaire response, and is zero otherwise. We use this participant representation later on for creating persona descriptors. 

To map a questionnaire response of $P_i$ into a feature vector $\bm{p}_i$, we need to perform annotation.  
For traits elicited from closed-ended codes, we did not require the annotators, since the participant's answer to a closed-ended question(s) could be directly mapped to a trait. However, for traits elicited from open-ended codes, we asked annotators to mark the absence/presence of a trait.

Nine annotators took part in the annotation process of 180 questionnaire responses. Concepts derived with an affinity diagram provided annotators with a clearer structure of the task and built connections between the traits, minimising annotators' fatigue. 
The annotation workload was split into eight parts: with respect to questionnaire responses, and the type of traits to be annotated. Since the generation set $\mathscr{G}$ consisted of a high number of responses, we partitioned it into three subsets. We did not split the validation set $\mathscr{V}$. We used the concepts derived from an affinity diagram to split the traits to be annotated. Annotators $A1$, $A2$, $A3$, $A4$ worked with traits related to the following affinity diagram concepts: beliefs about advertisement, monetisation, inference on participant's data, types of protective behaviour (prior to the first stimulus, after the first stimulus), beliefs about what the PET does, attitude to the PET, and beliefs about privacy online.  Annotators $A5$, $A6$, $A7$, $A8$ worked with traits related to concepts about the participants' level of knowledge, emotions after the first stimulus, categories of personal or private data, use of social media platforms, sensitivity towards possible privacy risks, and beliefs about images. Annotators $A1, ..., A8$ were assigned one of the eight parts of the annotation workload. Annotator $A9$ annotated all eight parts of the annotation workload. Each trait per participant was annotated by two annotators. The disagreements between the annotators were discussed during follow-up meetings and resolved in unanimous agreement. Based on the understanding of the annotation process in qualitative studies~\citep{elueze_privacy_2018, dupree_privacy_2016, stover_how_2023, xu_identifying_2020, braun_thematic_2022}, we do not compute Cohen's kappa~\citep{cohen_coefficient_1960} for the inter-annotator agreement.

\subsection{Likert-scale and Binary Variables}
While performing manual annotation, we noticed differences in the nature of the traits: some of the traits were mutually exclusive (e.g.,~if multiple traits related to the same closed-ended question, only one of those traits could be non-zero), whereas others were not (e.g.,~a participant could have expressed multiple traits in the same category, such as emotions). 
To incorporate these traits' differences in a dissimilarity measure between the participants, we first introduce the Likert-scale and binary explanatory variables. 
The Likert-scale explanatory variables, $l_{j}$, allow to group: 
\begin{enumerate}
    \item the open-ended traits that are mutually exclusive and measure the degree of certain activity/belief, e.g., "has shown to share little/moderate/a lot of personal information in the last 10 images"; 
    \item the closed-ended traits that come from the same closed-ended question, e.g., "they would never/.../always use the proposed PET"; 
    \item the closed-ended traits which are manually derived and measure the change in privacy protection importance, and control mismatch, e.g., "experienced a drastic decrease/ .../drastic increase in privacy protection importance".
\end{enumerate}
As a result, we obtain $L = 2 + 10 + 2 = 14$ Likert-scale explanatory variables. 
The binary explanatory variables, $b_i$, are propagated from the remaining open-ended traits, $B = 67$. We hence identify $E = L + B = 81$ explanatory variables based on the traits in $\mathcal{T}$. 
The full list of explanatory variables, their indices and corresponding traits is provided in Appendix \ref{app:explanatory_vars}.

We convert $\bm{p_i} \in \{0, 1\}^T$ to its alternative form: 
\begin{equation}
    \bm{p_i'} = \bm{l_i} \oplus \bm{b_i}, 
\end{equation}
where $\bm{l_i} \in \mathbb{R}^{L}$ is a vector of Likert-scale explanatory variables, $\bm{b_i}$ is the vector for binary explanatory variables, $\bm{b_i} \in \{0,1\}^B$; $\oplus$ is a vector concatenation operation.

The list of identified Likert-scale variables, $l_{*}~\in~[0,1]$, is provided below: 
\begin{enumerate}[label = $l_{\arabic*}$] 
    \item the amount of shared information, [a lot, few]
    \item the level of sensitivity to personal data, [low, high]
    \item the level of privacy protection importance at the beginning of the questionnaire, [unimportant, very important]
    \item the perception of how easy it is to protect privacy, [very~easy, very~hard]
    \item the desired level of control, [never, always]
    \item the level of awareness about social media checking the content, [not~..., very ...] aware
    \item the level of awareness about social media building profiles of users, [not~..., very~...] aware
    \item the level of understanding of what is happening to their data online, [no~..., full~...] understanding 
    \item the perceived level of control over their privacy, [no~..., full~...]~control
    \item the perception of PET's usefulness, [not useful, useful]
    \item how often they would use the proposed PET, [never, always]
    \item the level of privacy protection importance at the end of the questionnaire, [unimportant, very important] 
    \item the change in the privacy protection importance, [drastic decrease, drastic increase] 
    \item the mismatch between desired and perceived level of control over privacy, [extremely less~..., extremely more~...] control than wanted
\end{enumerate}

\section{Personas Elicitation}
\label{sec:dissimilarity_measure}

To elicit the personas we build on top of divisive hierarchical clustering. We propose a new dissimilarity measure, which incorporates the nature of open- and closed-ended questions by considering their corresponding explanatory variables. Additionally, we propose a two-step dendrogram pruning process, which uses Boschloo’s statistical test~\citep{boschloo_raised_1970} to validate differences between the clusters. 

\subsection{Dissimilarity Measure}

The closed-ended questions that are linked to Likert-scale explanatory variables require choosing between the set of possible pre-defined answers. For this kind of data norms like $L_1$ or $L_2$ are more suitable than a cosine similarity measure or a dot product\footnote{Dot product or cosine similarity metric should not be used in the case of Likert-scale questions. Treating Likert-scale questions as an unordered set of traits ignores the distances between the Likert items. If the Likert-scale variables are represented as our vector $\bm{l}$, cosine similarity between $\bm{l}_i$ and $\bm{l}_j$ can lead to participants with opposite views (e.g., $\bm{l}_i = (1,...,1)$, $\bm{l}_j = (5,...,5)$) being considered identical.}. 

Binary explanatory variables extracted from open-ended questions could not be predicted before the coding process -- the participant could not choose between the set of possible answers. Zero value for such a binary explanatory variable does not mean that a participant did not share the views of the trait formulation, it means that they did not express any opinion, rather than disagree with it. For example, the value of one for the trait "thinks that privacy online is deceptive" means that the participant has vocalised this view. However, a value of zero for this trait does not mean that the participant believes that privacy protection is transparent (non-deceptive). This means that using a metric like a dot product (like it was done in the work of \citet{dupree_privacy_2016}) is preferable over using $L_1$ metric for measuring dissimilarity, since $L_1$ penalises for each disagreement, while dot product focuses on the agreements only. However, sometimes an open-ended question can be associated with a Likert-scale variable (e.g., traits that describe the amount of information that a participant is sharing -- a lot, moderate, few). In such cases, $L_1$ or $L_2$ norms should be used. 

We merge the approaches\footnote{\citet{dupree_privacy_2016} have coded the survey responses which were of a closed-ended nature and performed hierarchical agglomerative clustering on a binary feature vector using the dot product. \citet{biselli_challenges_2022} have clustered participants using the hierarchical agglomerative clustering using Ward's linkage and $L_2$-norm.}  used by \citet{dupree_privacy_2016} and \citet{biselli_challenges_2022} by introducing a new dissimilarity measure, which takes into account the nature of the explanatory variable. To achieve this, we consider the Likert-scale part of a feature vector, $\bm{l}$, and the binary, $\bm{b}$, separately, since they are different in nature. To this end, we propose a new dissimilarity measure:
\begin{align}
    d(\bm{p_i'}, \bm{p_j'})  &= \max\Big(0, \ \frac{L_1(\bm{l_i}, \bm{l_j})}{\sum_{k = 1}^L r(l_k)} - \frac{\bm{b_i} \cdot \bm{b_j}}{B}\Big), 
    \label{eq:distances}
\end{align}
where $L_1$ is Manhattan distance, $r(\cdot)$ is a function for calculating the range of the $k$-th Likert explanatory variable, $l_k$. 
The maximum possible distance between the participants is defined by their Likert-scale responses normalised by the maximum distance in $L_1$. We adjust the distance obtained on $\bm{l}$ by a normalised dot product on $\bm{b}$. In Eq. (\ref{eq:distances}), $\max(0, \ \cdot)$ was added to avoid negative dissimilarity scores, which reduced the range of $d(\bm{p_i'}, \bm{p_j'})$ to $[0,1]$.

\subsection{Dendrogram Construction}
We use the dissimilarity measure to perform divisive hierarchical clustering. Hierarchical clustering offers a number of clusters at different levels of abstraction, starting with only one cluster for the whole population and ending with as many clusters as there are data points. We use divisive clustering and separate the most dissimilar instances at each level of a dendrogram~\citep{kaufman_finding_2005}. 
We denote a cluster $k$ at a level of granularity $v$ as $\mathcal{U}^v_k$. We represent a cluster $\mathcal{U}^v_k$ by using a cluster descriptor $\bm{u}^{v}_k$. The $t$-th entry of the descriptor is the frequency of appearance of $t$-th trait in a cluster, and is calculated as follows: 
\begin{equation}
    \bm{u}^{v(t)}_k = \frac{\sum_{\bm{p_i} \in \mathcal{U}^v_k}\bm{p_i}^{(t)}}{|\mathcal{U}^v_k|}.
\end{equation}

We now define a notation for the statistical similarity between clusters. By statistical similarity, we refer to clusters' descriptors having no differences based on a statistical measure. For this, we check if the frequency of appearance of the $t$-th trait is different between clusters using Boschloo's test~\citep{boschloo_raised_1970}, which is a test that is suitable for small datasets. 

Next, we perform discriminative feature selection. Once the initial dendrogram is built, we identify the most discriminative traits for the final dendrogram construction by reducing the number of traits elicited on the open-ended questions. We perform pairwise comparisons of the clusters in the first 15 levels of the dendrogram. We retain the trait $t_i$ if for at least one comparison the trait had  $p<0.001$. If a trait is associated with a binary explanatory variable and has a $p\ge 0.001$, it is removed. For Likert variables that consist of traits from open-ended questions, if at least one trait is significant, all of the traits of the corresponding variable are retained. This has reduced the number of traits that are significant from $T$ to $S = 72$. All other traits were masked out (set to zero) from the feature vectors $\bm{p_i}$ and $\bm{p_i'}$. We built a final dendrogram using only the discriminative features. 

\subsection{Pruning}

\begin{figure}[t!]
    \centering
    \includegraphics[width = 0.95\linewidth]{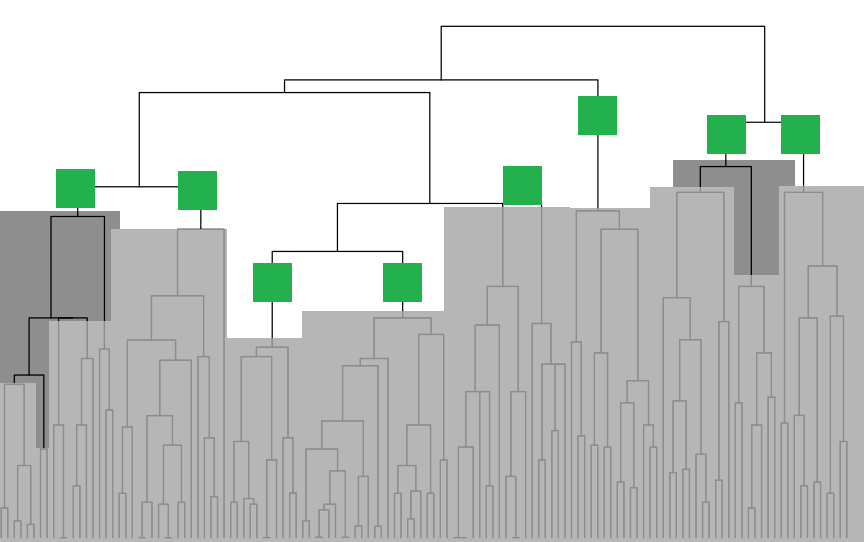}
    \caption{Our two-step method for discovering privacy personas. In step~one~{\fboxsep=0pt\fbox{\color{oh_color_gray_light}\rule{3mm}{3mm}}}, the dendrogram is pruned if a parent cluster is split into two sub-clusters that are statistically similar to each other, meaning there are no traits that make these clusters different based on Boschloo's test~\citep{boschloo_raised_1970}. In step~two~{\fboxsep=0pt\fbox{\color{oh_color_gray_dark}\rule{3mm}{3mm}}}, the dendrogram is further pruned if there exists at least one leaf that is statistically similar to other leaves. The final personas are in green~{\fboxsep=0pt\fbox{\color{oh_color_green_leaf}\rule{3mm}{3mm}}}.}
    \label{fig:step_1_2}
\end{figure}

Once we build the dendrogram with discriminative features, we prune it to avoid clusters being split into sub-clusters that are statistically similar. To compare clusters $\mathcal{U}^v_k$ and $\mathcal{U}^n_m$ we run Boschloo's test~\citep{boschloo_raised_1970} on the discriminative traits, running $S$ statistical tests per comparison of two clusters. We consider clusters $\mathcal{U}^v_k$ and $\mathcal{U}^n_m$ to be significantly different if there exists at least one trait for which the difference in proportions\footnote{We scale 0.05 by $S$, $S-1$, $S-2$... to allow for Holm-Bonferroni correction.} is $p<0.05$. 
 
We now follow a two-step process for pruning the dendrogram:~\begin{enumerate}
    \item each cluster $\mathcal{U}^v_k$ is split into sub-clusters $\mathcal{U}^{v+1}_m$ and $\mathcal{U}^{v+1}_n$ if and only if there exists at least one trait of significance $p<0.05$; otherwise we say that cluster $\mathcal{U}^v_k$ is non-divisible. 
    \item leaves of the dendrogram are merged into their parent cluster if they are statistically similar to other leaves. For each leaf, we perform a comparison with other leaves and count the number of comparisons for which the leaves are not statistically different. We select a cluster with the highest number of insignificant comparisons and go one step back in the dendrogram, merging the leaf with the highest number of insignificant comparisons and its sibling(s) into their parent cluster. We iteratively repeat this process until for pairwise comparisons of all of the leaves in a tree there exists at least one significantly different trait. 
\end{enumerate}

After following Step 1 we obtain twelve leaves, which after Step~2 got reduced to eight leaves (see Fig. \ref{fig:step_1_2}). We call the obtained leaves the privacy personas. This process allows us to elicit privacy personas that are different to each other based on a statistical measure. Additionally, we compared the confidence intervals (CI) on the traits of the personas, checking if 95\% CI for the traits would overlap~\citep{agresti_approximate_1998}. For all comparisons, there exists at least one trait for which confidence intervals are not overlapping, supporting the differences between the personas.

\section{Privacy Personas}
\label{sec:privacy_personas}

\begin{figure*}
    \centering
    \includegraphics[width=0.95\linewidth]{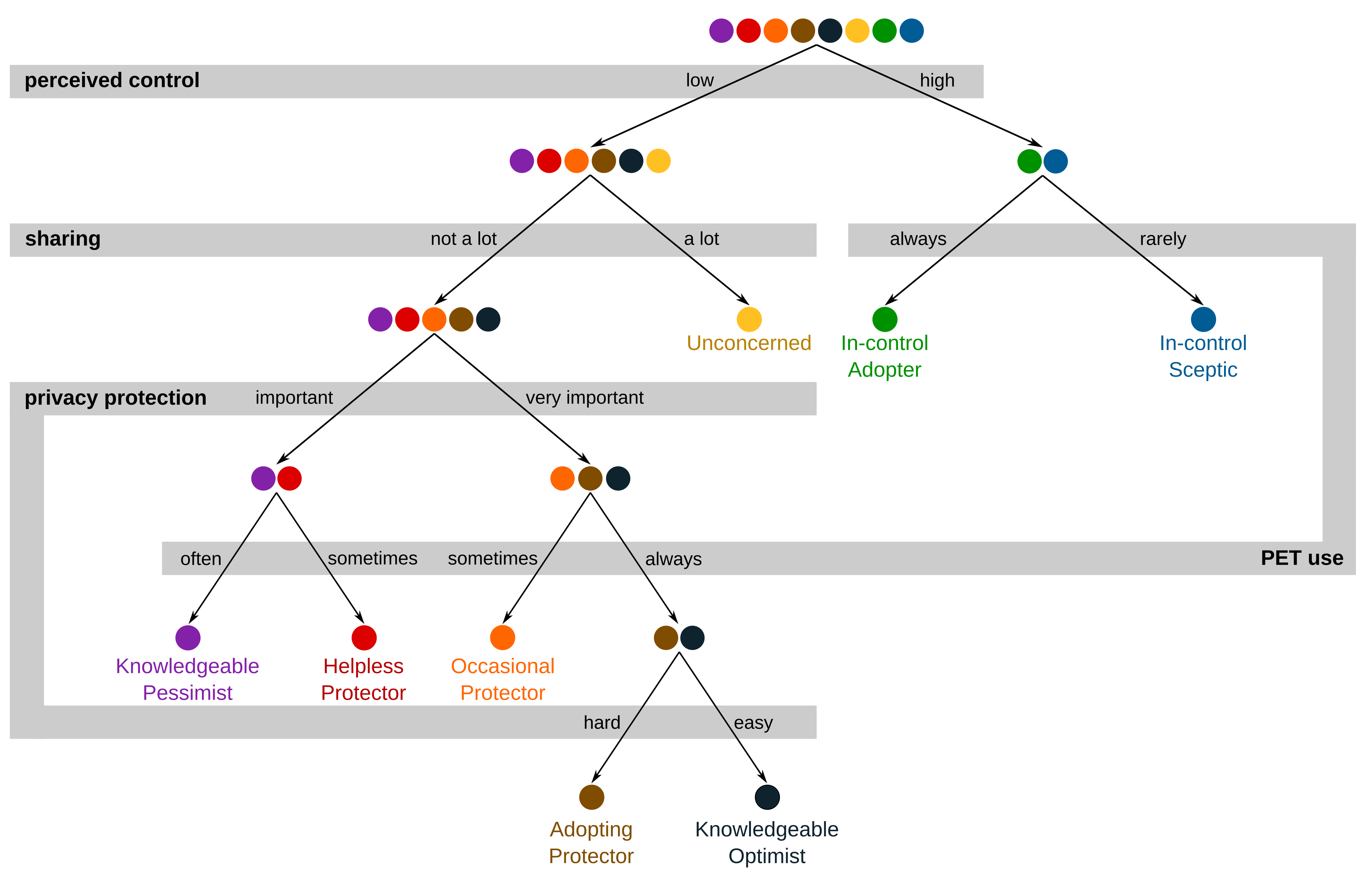}
    \caption{The discriminative features of our identified privacy personas. Text in bold defines an attribute on which the personas differ from each other, the values next to the arrows are the values of a corresponding attribute.}
    \label{fig:decision_tree}
\end{figure*}

The hierarchical structure identified in Section \ref{sec:dissimilarity_measure} allows us to define eight privacy personas as leaf clusters (see Fig. \ref{fig:decision_tree}), which are described below. 

\textbf{Unconcerned}. 
This persona shares the most personal information online out of all personas. They have the lowest desired level of control and have no opinion on the importance of privacy protection. They have a significant level of mismatch between the perceived and desired level of control, and would rarely use PET.

\textbf{In-control Adopter}. This persona has a high level of perceived control and they have a strong faith in the usefulness of PET. This persona has only a slight mismatch between the desired level of control and the perceived one. They would almost always use the proposed PET. 

\textbf{In-control Sceptic}. This persona is the most sceptical about the proposed PET while having one of the highest levels of perceived control. 

\textbf{Helpless Protector}. This persona has only a slight understanding of what is happening to their data. They think that the PET should somehow help to protect their data, but would only occasionally use it.

\textbf{Knowledgeable Pessimist}. For this persona, privacy protection is very hard and they believe that the PET should somehow help to protect their data, and they would use it. 

\textbf{Occasional Protector}. This persona sees privacy protection as a very important task. While this persona always wants to be in control over their data, they would only sometimes use the PET. 

\textbf{Adopting Protector}. This persona wants to always be in control of their data and sees privacy protection as a very important, but hard task. They are convinced that the PET would allow them to protect their privacy and would always use it. 

\textbf{Knowledgeable Optimist}. This is the only persona who strongly believes that privacy protection is mostly easy. This is the second most knowledgeable persona. They see privacy as very important, and would always want to be in control of their data. Because of their knowledge, they see privacy as a mostly easy task. They have a strong faith in the PET and would almost always use it. 

The persona cluster sizes are: \textit{Unconcerned} -- 14, \textit{In-control Adopter} -- 18, \textit{In-control Sceptic} -- 11, \textit{Helpless Protector} -- 17, \textit{Knowledgeable Pessimist} -- 18, \textit{Occasional Protector} -- 18, \textit{Adopting Protector}~-- 11, and \textit{Knowledgeable Optimist} -- 23.

We discuss our personas in more detail in Section \ref{sec:other_privacy_personas}, where we analyse them with respect to their perception of privacy protection importance (comparing to Westin~\citep{ponnurangam__kumaraguru_privacy_2005}), their knowledge and behaviour (comparing to \citet{biselli_challenges_2022}), their motivation and knowledge (comparing to \citet{dupree_privacy_2016}), and their behaviour and privacy protection importance perception (comparing to~\citet{schomakers_typology_2019}). More detailed descriptors of the personas are provided in Appendices \ref{app:descriptors}, \ref{app:likert_desc}, and \ref{app:2D_spaces}.

\section{Privacy Personas Validation} 
\label{sec:validation}
In this section, we test the sensitivity of the dendrogram to small perturbations in the dataset (when a subset of participants from a generation set is removed), and participants' saturation (a measure of similarity of participants in the validation and generation sets).

\subsection{Sensitivity Analysis}

We evaluate the sensitivity of the final clustering structure to the changes in the dataset. To this end, we use the Fowlkes-Mallows (FM)  Index~\cite{fowlkes_method_1983}, which allows us to compare two hierarchical clustering results by incorporating information about their topology and label assignment~\cite{fowlkes_method_1983}. 
FM takes cluster assignment of a set of the same datapoints as an input. Hence, for computing the FM Index (the higher the value, the better the clustering), we compare the dendrogram obtained on a full generation set $\mathscr{G}$ ($n= G$ datapoints) and a dendrogram obtained on a reduced dataset ($n - r$ datapoints): (1) we consider both dendrograms at a level of granularity $v$; (2)~remove the corresponding $r$ datapoints from the labels list obtained on a full generation set $\mathscr{G}$; (3) compute the FM Index.

We compute the FM Index for $r = 1,2,..., 6$, selecting half of the minimal persona cluster size as a threshold for $r$. To ensure the validity of the index, we compute it 500 times, each time randomly sampling $n - r$ datapoints from $\mathscr{G}$. See Fig. \ref{fig:mean_FM} for the mean of the distributions for different $r$ and Fig. \ref{fig:FM_dist} in Appendix \ref{app:sensitivity_analysis} for more detailed results.
For all $r$ when $v = 3$ the $\overline{FM} \geq 0.8$, which means that at least to a level of granularity 3 our results are stable. For all $r$ we see a trend of $\overline{FM}$ gradually decreasing, where the $\overline{FM}$ starts to drop significantly for all $r$ at $v = 10$. The splits in a dendrogram at $v = 2,3$ are the most stable ones, whereas after $v = 10$ the dendrogram becomes more sensitive to noise. We attribute the noise to the rejected cluster splits, rejecting the first cluster split at a level of granularity $v = 5$. The last accepted split of a parent cluster was a level $v = 15$, which is just above the 0.6 threshold for most $r$ values. Even with the rejected clusters the FM Index stays high, supporting the stability of our clustering. 

\subsection{Validation of Participants' Saturation}
To validate that we have reached participants' saturation, we compare the annotated questionnaires from the $\mathscr{G}$ and $\mathscr{V}$, which were annotated by $A1$, ..., $A7$, $A9$ and $A4$, $A8$, $A9$, respectively. We pose the following question: are there any participants from $\mathscr{V}$ that are considered to be outliers with respect to the distribution of participants in $\mathscr{G}$? 

\begin{figure}[t]
    \centering
    \includegraphics[width = 0.97\linewidth]{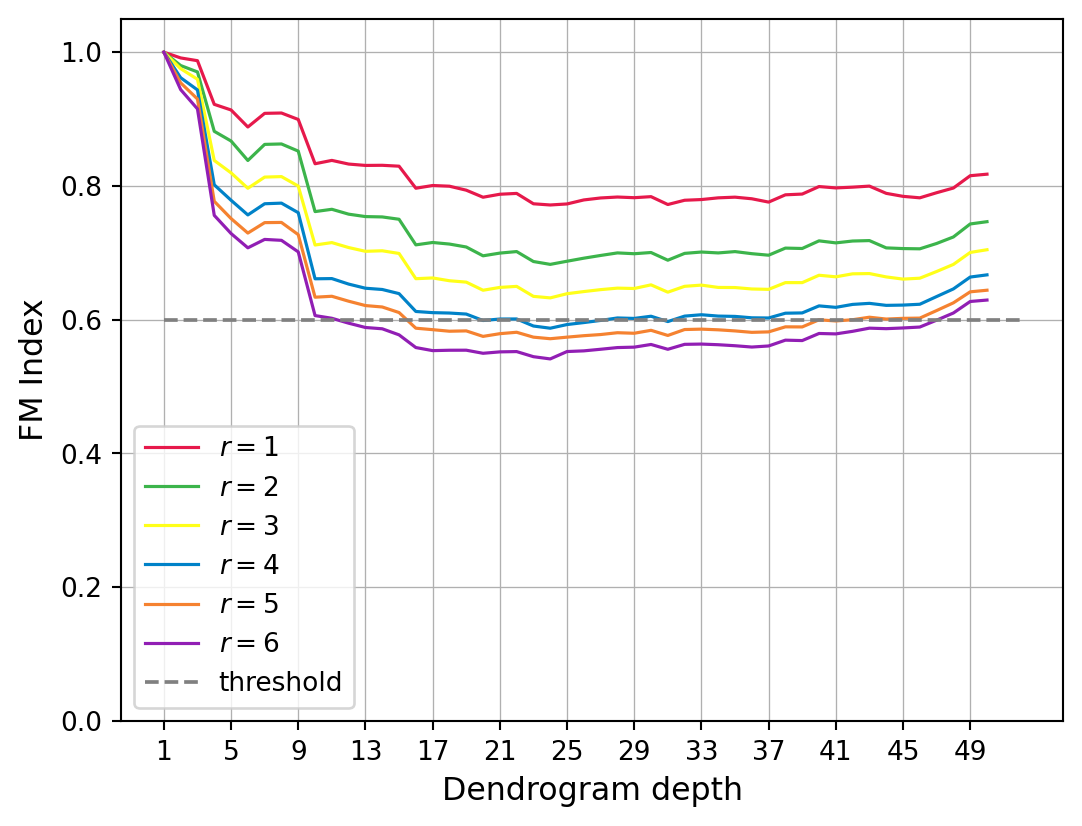}
    \caption{Sensitivity analysis of the dendrogram obtained on the generation set $\mathscr{G}$. We randomly remove $r$ participants from the generation set $\mathscr{G}$, form a new dendrogram, and compute the Fowlkes-Mallows (FM) Index~\citep{fowlkes_method_1983} between the newly obtained dendrogram and a dendrogram obtained on the initial generation set $\mathscr{G}$. We repeat the sampling procedure 500 times for each value of $r$, and report the mean value. We notice that the highest drop in performance takes place at a dendrogram depth equal to 10 for all $r$. For more detailed plots of the sensitivity analysis, see Appendix \ref{app:sensitivity_analysis}.}
    \label{fig:mean_FM}
\end{figure}

\begin{figure*}[ht!]
    \centering
    \includegraphics[width = 0.95\linewidth]{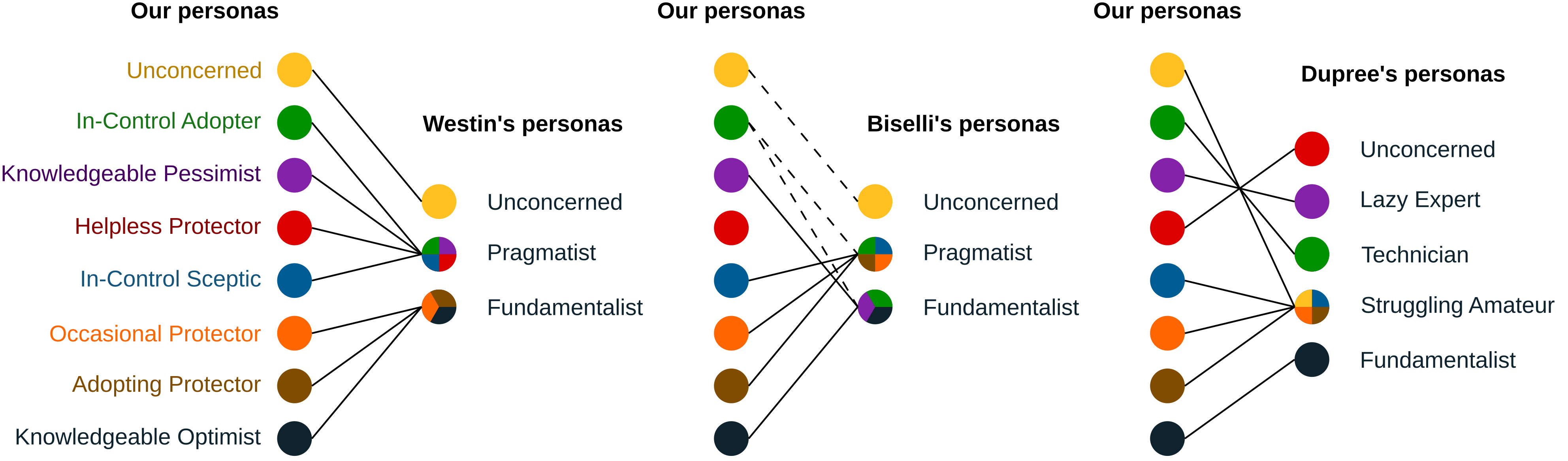}
    \caption{Mapping of our personas into Westin's~\citep{ponnurangam__kumaraguru_privacy_2005} (left), Biselli's~\citep{biselli_challenges_2022} (centre), and Dupree's~\citep{dupree_privacy_2016} (right). Our personas are mapped into Westin's by considering the privacy protection importance attribute: high, moderate and low privacy protection importance is mapped into Westin's Fundamentalist, Pragmatist and Unconcerned personas accordingly. Our personas are mapped into Biselli's by considering knowledge and behaviour attributes: high, moderate and low knowledge and privacy-protective behaviour are mapped into Biselli's Fundamentalist, Pragmatist and Unconcerned personas accordingly). Our personas are mapped into Dupree's by considering primarily the knowledge attribute.}
    \label{fig:westin_biselli_dupree_ours_personas}
\end{figure*}

Both coders $C1$, and $C2$ stated that saturation was reached after both of them coded a corresponding set of questionnaire responses. Based on this, we make the following assumption: within our generation set $\mathscr{G}$, each of the participants has at least one person to whom they are similar. 

Since the validation set $\mathscr{V}$ is smaller than the generation set $\mathscr{G}$, we do not aim to have a direct comparison of the distributions of participants in $\mathscr{G}$ and $\mathscr{V}$. Instead, we first obtain the distribution of the distances to the closest neighbour for all $P_i \in \mathscr{G}$. Then for each $P_j \in \mathscr{V}$, we find the nearest neighbour from $\mathscr{G}$. We check if the corresponding distance is an outlier in the distribution of the distances to the closest neighbour for all $P_i \in \mathscr{G}$.

To achieve this we compute two distance matrices:
\begin{itemize}
    \item $\bm{D_1} = [d_{ij}]$, where $i,j = 1, 2, ..., 130$; $d_{ij} = d(\bm{p_i'}, \bm{p_j'})$ if $i \neq j$, $d_{ij} = 1$ otherwise (to avoid self-similarities); $\bm{p'_i} , \bm{p'_j} \in \mathscr{G}$.  
    \item $\bm{D_2} = [d_{pq}]$, where $p = 1, 2, ..., 130$, $q = 1, 2, ..., 50$; $d_{pq} =d(\bm{p'_p}, \bm{p'_q}) \ \forall p, \ q$; $\bm{p'_p} \in \mathscr{G}$  and $\bm{p'_q} \in \mathscr{V}$.
\end{itemize}

We then calculate two vectors, $\bm{d_1}$ and $\bm{d_2}$: 
\begin{equation}
    \bm{d_1} = [d_{\cdot j}] = [\min_{i = 1}^{130} d_{ij}] \quad \textrm{and} \quad \bm{d_2} = [d_{\cdot q}] = [\min_{p = 1}^{130} d_{pq}], 
\end{equation}
where $\bm{d_1}$ contains the distances to the closest neighbour in $\mathscr{G}$ for each $P_i \in \mathscr{G}$ and $\bm{d_2}$ contains the distances to the closest neighbour in $\mathscr{G}$ for each $P_j \in \mathscr{V}$. Based on Tukey's method~\citep{tukey_exploratory_1977}, none of the elements in $\bm{d_2}$ are outliers, since all of the $z$-scores for elements in $\bm{d_2}$ are within $[-1.25, 2.05]$ range. This means that for each participant in $\mathscr{V}$, there exists at least one participant in $\mathscr{G}$ to which they are similar. 

\section{Relationship with Other Personas}
\label{sec:other_privacy_personas}

In this section, we compare our personas and the privacy personas identified by Westin~\citep{ponnurangam__kumaraguru_privacy_2005}, \citet{biselli_challenges_2022}, \citet{dupree_privacy_2016} and \citet{schomakers_typology_2019}. For comparison with the related work, we selected the explanatory variables that best reflect the underlying attribute used to define personas.

\subsection{Westin}
\label{subsec:westin_vs_ours}

The main classification criteria used by Westin~\citep{ponnurangam__kumaraguru_privacy_2005} was the level of participants' concern about privacy, which can be mapped to privacy protection importance. By considering traits that capture only privacy protection importance, we can map our personas into Westin's (see Fig. \ref{fig:westin_biselli_dupree_ours_personas}).
Our personas refine Westins' by considering additional privacy attributes. Our personas support the findings of \citet{king_taken_2014} that Westin's personas classification does not help to predict "knowledge, behaviours, or an alternative measure of attitudes"~\citep{king_taken_2014}. The reason for this is that questions about privacy protection importance are not rich enough to capture the space of privacy personas, losing variability that corresponds to knowledge and behaviour. The level of privacy protection importance can only be informative when defining the \textit{Unconcerned} persona. 

\citet{urban_privacy_2014} considered Westin's classification in a different light. They proposed to consider Westin's Fundamentalist as a privacy-resilient persona, and the Pragmatist and the Unconcerned as pri\-vacy-vulnerable personas. They hypothesised that the privacy-resilient persona would be more willing to protect their privacy, whereas privacy-vulnerable ones would be more hesitant to use self-help technologies. We use our personas as a proxy for validating this hypothesis. 

We observe that Westin's Unconcerned, which has a one-to-one mapping with our Unconcerned, is the least willing persona to use the proposed PET. However, when it comes to Westin's Pragmatist, we cannot predict how willing one would be to use the PET. We partitioned Westin's Pragmatist with four of our personas, namely \textit{In-Control Adopter}, \textit{Knowledgeable Pessimist} (both are willing to use the PET), \textit{Helpless Protector} (somehow willing to the use PET), and \textit{In-Control Sceptic} (not willing to use PET). Similarly, we partitioned Westin's Fundamentalist with three of our personas, namely \textit{Adopting Protector} and \textit{Knowledgeable Optimist} (both are willing to use the PET), and \textit{Occasional Protector} (who is somehow willing to use the proposed PET). While the least willing personas come from Westin's Unconcerned and Pragmatist clusters, some of Westin's Pragmatists are willing to use the PET. Similarly, Westin's Fundamentalists could have a different degree of willingness to use PET. This means that it is not sufficient to use Westin's categorisation to predict if one would use the PET. 

\subsection{Biselli}
\label{subsec:biselli_vs_ours}

\begin{figure}[ht!]
    \centering
    \includegraphics[width= 0.77\linewidth]{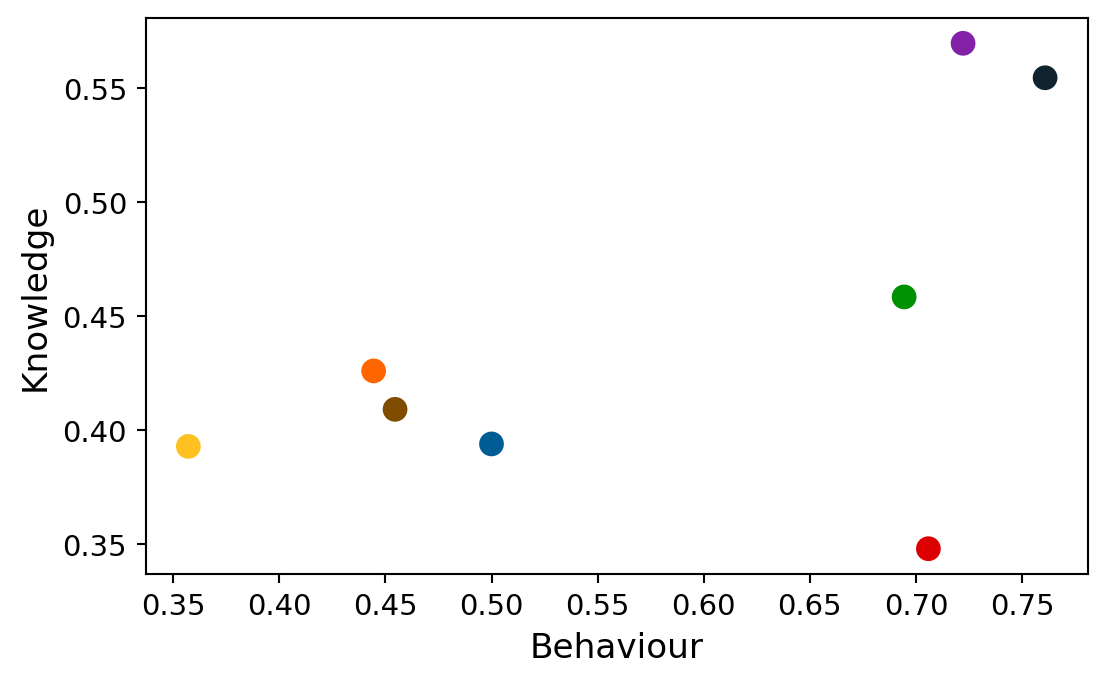}
    \includegraphics[width= 0.77\linewidth]{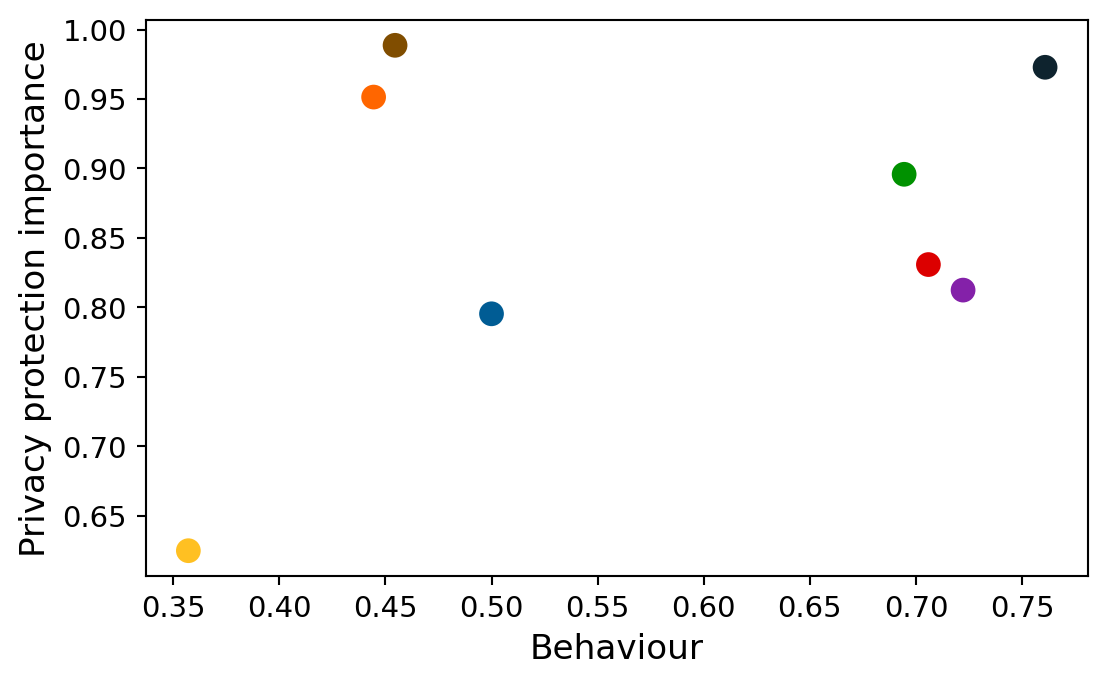}
    \caption{Projection of our personas into 2D spaces: behaviour-knowledge space (top), behaviour-privacy protection importance (bottom). Zero/one for behaviour corresponds to sharing few/a lot of personal elements in their images (maps to $l_1$ explanatory variable). Zero/one for knowledge is low/high level of knowledge (maps to $(l_6 + l_7 + l_8)/3$). Zero/one for privacy protection importance is low/high level of importance (maps to $(l_3 + l_{12})/2$).  Colour-coding: Unconcerned (\tikzcircle[oh_yellow, fill=oh_yellow]{3pt}), 
    In-control Adopter (\tikzcircle[oh_green, fill=oh_green]{3pt}), 
    In-Control Sceptic (\tikzcircle[oh_blue, fill=oh_blue]{3pt}), 
    Knowledgeable Pessimist (\tikzcircle[oh_purple, fill=oh_purple]{3pt}), 
    Helpless Protector (\tikzcircle[oh_red, fill=oh_red]{3pt}), 
    Occasional Protector (\tikzcircle[oh_orange, fill=oh_orange]{3pt}), 
    Adopting Protector (\tikzcircle[oh_brown, fill=oh_brown]{3pt}), 
    Knowledgeable Optimist (\tikzcircle[oh_black, fill=oh_black]{3pt}).
    }
    \label{fig:2d_proj_be_kn_ppi}
\end{figure}

\citet{biselli_challenges_2022} elicited personas using only two attributes, \textit{behaviour} and \textit{knowledge}, and showed a strong positive correlation between these. They elicited three personas, adopting Westin's~\citep{ponnurangam__kumaraguru_privacy_2005} naming convention: the Fundamentalist (high knowledge, high privacy-enhancing behaviour), the Pragmatist (moderate knowledge, moderate privacy-enhancing behaviour), and the Unconcerned (low knowledge, low privacy-enhancing behaviour). 
To compare our personas and Biselli's, we first project our identified personas into the traits with the attributes used by Biselli et al.~\citep{biselli_challenges_2022}: \textit{knowledge} and \textit{behaviour} (see Fig. \ref{fig:2d_proj_be_kn_ppi}). 

We observe that some of our personas obey the positive correlation discovered by \citet{biselli_challenges_2022}. For example, \textit{Knowledgeable Pessimist} and \textit{Knowledgeable Optimist} both fall under the category of high knowledge and high privacy-preserving behaviour. \textit{Occasional Protector}, \textit{Adopting Protector} and \textit{In-Control Sceptic} fall under the category of medium knowledge and medium privacy-preserving behaviour. The only persona that shows the lowest level of protective behaviour is our \textit{Unconcerned}, however, this persona does not show the lowest level of knowledge. We therefore say that our \textit{Unconcerned} can only be weakly mapped to Biselli's. Similarly, our \textit{In-control Adopter} can be weakly mapped to both Biselli's Fundamentalist and Pragmatist, since they both have a high level of privacy-protecting behaviour and medium knowledge. We also identify a new type of persona: \textit{Helpless Protector}, which has the lowest level of knowledge, but a high level of protective behaviour. 

 While some of our personas follow similar patterns as personas elicited by \citet{biselli_challenges_2022}, they allow for more granularity. For example, by considering only \textit{knowledge} and \textit{behaviour}, the difference between our \textit{Knowledgeable Optimist} and \textit{Knowledgeable Pessimist} would have been lost (e.g., the perception of how easy it is to protect privacy).

\subsection{Dupree}
\label{subsec:dupree_vs_ours}
\citet{dupree_privacy_2016} elicited five personas using a mixture of qualitative and quantitative analysis. They mapped their Fundamentalist to Westin's Fundamentalist, their Marginally Concerned to Westin's Unconcerned, and the remaining personas (Lazy Expert, Struggling Amateur, and Technician) were segmenting Westin's Pragmatic majority. \citet{dupree_privacy_2016} positioned their personas in the knowledge-motivation space. Since \citet{ferrarello_reframing_2022}  were measuring knowledge, but not motivation, we use the level of knowledge, as the primary factor for mapping. \citet{dupree_privacy_2016} identified four levels of knowledge among their personas: low (Moderately Concerned), medium-low (Struggling Amateur), medium-high (Technician), and high (Fundamentalist and Lazy Expert). We map our personas into these categories based on their knowledge, using the four levels of knowledge for this assignment. 

Our \textit{Knowledgeable Optimist} and \textit{Knowledgeable Pessimist} have the highest level of knowledge and can be mapped to either Dupree's Fundamentalist or Lazy Expert (see Fig. \ref{fig:westin_biselli_dupree_ours_personas}). To disentangle this mapping, one would need to record the future behaviour of these personas. However, given that \textit{Knowledgeable Pessimist} wants to be in control over their data often (and not always), this could mean that they are closer to a Dupree's Lazy Expert than to a Fundamentalist. Similarly, based on the level of knowledge only, our \textit{In-control Adopter} could be mapped to Dupree's Technician. Then, Dupree's Struggling Amateurs can be represented by four of our personas: \textit{Unconcerned, In-control Sceptic, Occasional Protector}, and \textit{Adopting Protector}, given that these personas have similar levels of knowledge. Finally, our \textit{Helpless Protector} has the lowest level of knowledge, making them Dupree's Unconcerned. 

While we approximate the mapping by the level of personas' knowledge, we do not have a direct measurement of a motivation attribute. The closest of our attributes are motivation to use a PET,  perception of privacy importance, and desired level of control, however, none of these has a direct link with Dupree's motivation. 

\subsection{Schomakers}
\label{subsec:schomakers_vs_ours}

Once \citet{schomakers_typology_2019} established the clusters, they were analysed with respect to participants' level of awareness, experience, perceived privacy, need for privacy, trust in online companies, and privacy self-efficacy. Notably, these attributes were not used during the clustering process. This means that even if there was a level of variability among these attributes, it was not captured. Values for these attributes could be averaging across multiple hidden sub-personas, hence should not be used for a comparison. This reduces the list of attributes for comparison to concern and behaviour. The closest attributes in our personas are the behaviour and the privacy protection importance (see Fig. \ref{fig:2d_proj_be_kn_ppi}). Based on this, we can conclude that our \textit{Knowledgeable Optimist} can be mapped to Schomakers' Guardian, having a high level of concern and protective behaviour. Similarly, \textit{Occasional Protector} and \textit{Adopting Protector}, could be mapped to Schomakers' Cynic, having a high level of privacy protection importance, but a lower level of demonstrated protective behaviour. Our \textit{In-Control Adopter}, \textit{Helpless Protector, Knowledgeable Pessimist}, and \textit{In-Control Sceptic} can be mapped to the Pragmatists majority. Finally, our method also provides an additional persona~-- \textit{Unconcerned}, which is not represented by~\citet{schomakers_typology_2019}.

\section{Limitations}
\label{sec:limitations}

The dataset~\citep{ferrarello_reframing_2022} showcased the responses from the participants from the UK, however, the cultural differences also influence privacy perception~\citep{xu_dipa2_2023}. There are conflicting opinions if age influences~\citep{biselli_challenges_2022, zhou_online_2017}, or does not influence~\citep{goyeneche_linked_2024} privacy perception, however, it could be one of the influencing factors in privacy personas. We perform validation on a subset of the same dataset, and not a new dataset. All of this means that while these elicited personas are valid for this dataset, additional validation with people from other countries and other age groups is beneficial. People who do not use social media were not represented in the study by \citet{ferrarello_reframing_2022}, and they might represent other personas (i.e. highly secure ones~\citep{dupree_privacy_2016}). We plan to collect more data and use the approach by \citet{caines_automatically_2018} for automating the annotation.

We did not record the user interactions with an application, as it is done in behavioural data studies~\citep{wisniewski_profiling_2014, mugan_understandable_2011}. Additionally, measuring user preferences towards the protection of a specific type of content~\citep{milne_information_2017, lin_modeling_2014} was out of the scope of our paper. Bridging the gap between modelling users by considering their privacy attitudes and perceptions (via surveys), factual behaviours (via user's interactions with applications), the type of content to be protected and its use is a part of our future work. 

\section{Conclusion}
\label{sec:conclusion}

We proposed a method that combines qualitative and quantitative approaches for discovering statistically different privacy personas. We used open coding and Boschloo's test~\citep{boschloo_raised_1970}, and accounted for the nature of the people's responses to questions (closed-ended vs open-ended). We built a dendrogram structure of clusters of participants, performed a two-step pruning process to ensure that the elicited personas were different to each other, and introduced a new measure for calculating this difference.  We discovered eight personas: \textit{Knowledgeable Optimist, In-control Adopter, In-Control Sceptic, Knowledgeable Pessimist, Helpless Protector, Occasional Protector, Adopting Protector, and
Unconcerned.} We provided personas' descriptors that allow one to understand the persona's behaviour, attitudes and other attributes, and have a numerical comparison between the personas. We have shown that it is important to consider the attitude towards privacy protection importance, participants' behaviours, decisions, and knowledge when discovering personas. Considering these attributes together when eliciting personas allows one to capture more complex patterns, which leads to better persona understanding, and hence more fine-grained privacy support. 

Relevant use cases of our personas include facilitating persona-tailored privacy support, accounting for the privacy needs of each persona according to their descriptor, modelling user interactions, and identifying privacy threats that can be posed in an application. Finally, privacy personas can be used for recruiting people for privacy studies based on their persona type, and not demographics only. 
In future work, we will create a larger dataset and explore automatic trait detection to allow for scalability and its applicability to other domains. We will also minimise the number of questions that could be used to elicit personas to have a short and reliable questionnaire by validating the scales using reliability and validity measures~\citep{novick_coefficient_1967, cho_making_2016}. Finally, we aim to use the identified privacy personas to support the design of online privacy settings~\citep{stover_how_2023, dupree_privacy_2016, biselli_challenges_2022}.

\begin{acks}
We thank the anonymous reviewers and the revision editor for their detailed comments, which helped improve our work.  This research received no specific grant from any funding agency in the public, commercial, or not-for-profit sectors. 
\end{acks}

\bibliographystyle{ACM-Reference-Format}
\bibliography{references}


\begin{thebibliography}{41}


\ifx \showCODEN    \undefined \def \showCODEN     #1{\unskip}     \fi
\ifx \showDOI      \undefined \def \showDOI       #1{#1}\fi
\ifx \showISBNx    \undefined \def \showISBNx     #1{\unskip}     \fi
\ifx \showISBNxiii \undefined \def \showISBNxiii  #1{\unskip}     \fi
\ifx \showISSN     \undefined \def \showISSN      #1{\unskip}     \fi
\ifx \showLCCN     \undefined \def \showLCCN      #1{\unskip}     \fi
\ifx \shownote     \undefined \def \shownote      #1{#1}          \fi
\ifx \showarticletitle \undefined \def \showarticletitle #1{#1}   \fi
\ifx \showURL      \undefined \def \showURL       {\relax}        \fi
\providecommand\bibfield[2]{#2}
\providecommand\bibinfo[2]{#2}
\providecommand\natexlab[1]{#1}
\providecommand\showeprint[2][]{arXiv:#2}

\bibitem[Agresti and Coull(1998)]%
        {agresti_approximate_1998}
\bibfield{author}{\bibinfo{person}{Alan Agresti} {and} \bibinfo{person}{Brent~A. Coull}.} \bibinfo{year}{1998}\natexlab{}.
\newblock \showarticletitle{Approximate {Is} {Better} than "{Exact}" for {Interval} {Estimation} of {Binomial} {Proportions}}.
\newblock \bibinfo{journal}{\emph{The American Statistician}} \bibinfo{volume}{52}, \bibinfo{number}{2} (\bibinfo{date}{May} \bibinfo{year}{1998}), \bibinfo{pages}{119--126}.
\newblock
\urldef\tempurl%
\url{https://doi.org/10.2307/2685469}
\showDOI{\tempurl}


\bibitem[Biselli et~al\mbox{.}(2022)]%
        {biselli_challenges_2022}
\bibfield{author}{\bibinfo{person}{Tom Biselli}, \bibinfo{person}{Enno Steinbrink}, \bibinfo{person}{Franziska Herbert}, \bibinfo{person}{Gina~M. Schmidbauer-Wolf}, {and} \bibinfo{person}{Christian Reuter}.} \bibinfo{year}{2022}\natexlab{}.
\newblock \showarticletitle{On the {Challenges} of {Developing} a {Concise} {Questionnaire} to {Identify} {Privacy} {Personas}}.
\newblock \bibinfo{journal}{\emph{Proceedings on Privacy Enhancing Technologies}} \bibinfo{number}{4} (\bibinfo{date}{Oct.} \bibinfo{year}{2022}), \bibinfo{pages}{645--669}.
\newblock
\urldef\tempurl%
\url{https://doi.org/10.56553/popets-2022-0126}
\showDOI{\tempurl}


\bibitem[Boschloo(1970)]%
        {boschloo_raised_1970}
\bibfield{author}{\bibinfo{person}{Ronald~D. Boschloo}.} \bibinfo{year}{1970}\natexlab{}.
\newblock \showarticletitle{Raised {Conditional} {Level} of {Significance} for the 2×2‐{Table} {When} {Testing} the {Equality} of {Two} {Probabilities}}.
\newblock \bibinfo{journal}{\emph{Statistica Neerlandica}} \bibinfo{volume}{24}, \bibinfo{number}{1} (\bibinfo{date}{March} \bibinfo{year}{1970}).
\newblock
\urldef\tempurl%
\url{https://doi.org/10.1111/j.1467-9574.1970.tb00104.x}
\showDOI{\tempurl}


\bibitem[Braun and Clarke(2022)]%
        {braun_thematic_2022}
\bibfield{author}{\bibinfo{person}{Virginia Braun} {and} \bibinfo{person}{Victoria Clarke}.} \bibinfo{year}{2022}\natexlab{}.
\newblock \bibinfo{booktitle}{\emph{Thematic {Analysis}: a {Practical} {Guide}}}.
\newblock \bibinfo{publisher}{SAGE Publications}, \bibinfo{address}{London, UK; Thousand Oaks, US}.
\newblock


\bibitem[Caines et~al\mbox{.}(2018)]%
        {caines_automatically_2018}
\bibfield{author}{\bibinfo{person}{Andrew Caines}, \bibinfo{person}{Sergio Pastrana}, \bibinfo{person}{Alice Hutchings}, {and} \bibinfo{person}{Paula~J. Buttery}.} \bibinfo{year}{2018}\natexlab{}.
\newblock \showarticletitle{Automatically {Identifying} the {Function} and {Intent} of {Posts} in {Underground} {Forums}}.
\newblock \bibinfo{journal}{\emph{Crime Science}} \bibinfo{volume}{7}, \bibinfo{number}{1} (\bibinfo{date}{Dec.} \bibinfo{year}{2018}).
\newblock
\urldef\tempurl%
\url{https://doi.org/10.1186/s40163-018-0094-4}
\showDOI{\tempurl}


\bibitem[Cho(2016)]%
        {cho_making_2016}
\bibfield{author}{\bibinfo{person}{Eunseong Cho}.} \bibinfo{year}{2016}\natexlab{}.
\newblock \showarticletitle{Making {Reliability} {Reliable}: a {Systematic} {Approach} to {Reliability} {Coefficients}}.
\newblock \bibinfo{journal}{\emph{Organizational Research Methods}} \bibinfo{volume}{19}, \bibinfo{number}{4} (\bibinfo{date}{Oct.} \bibinfo{year}{2016}), \bibinfo{pages}{651--682}.
\newblock
\urldef\tempurl%
\url{https://doi.org/10.1177/1094428116656239}
\showDOI{\tempurl}


\bibitem[Cohen(1960)]%
        {cohen_coefficient_1960}
\bibfield{author}{\bibinfo{person}{Jacob Cohen}.} \bibinfo{year}{1960}\natexlab{}.
\newblock \showarticletitle{A {Coefficient} of {Agreement} for {Nominal} {Scales}}.
\newblock \bibinfo{journal}{\emph{Educational and Psychological Measurement}} \bibinfo{volume}{20}, \bibinfo{number}{1} (\bibinfo{date}{April} \bibinfo{year}{1960}), \bibinfo{pages}{37--46}.
\newblock
\urldef\tempurl%
\url{https://doi.org/10.1177/001316446002000104}
\showDOI{\tempurl}


\bibitem[Colnago et~al\mbox{.}(2022)]%
        {colnago_is_2022}
\bibfield{author}{\bibinfo{person}{Jessica Colnago}, \bibinfo{person}{Lorrie~F. Cranor}, \bibinfo{person}{Alessandro Acquisti}, {and} \bibinfo{person}{Kate~H. Jain}.} \bibinfo{year}{2022}\natexlab{}.
\newblock \showarticletitle{Is {It} a {Concern} or a {Preference}? {An} {Investigation} into the {Ability} of {Privacy} {Scales} to {Capture} and {Distinguish} {Granular} {Privacy} {Constructs}}. In \bibinfo{booktitle}{\emph{Proceedings of the {Eighteenth} {USENIX} {Conference} on {Usable} {Privacy} and {Security}}} \emph{(\bibinfo{series}{{SOUPS} 2022})}. \bibinfo{publisher}{USENIX Association}, \bibinfo{address}{Boston, US}.
\newblock


\bibitem[Corbin and Strauss(1990)]%
        {corbin_grounded_1990}
\bibfield{author}{\bibinfo{person}{Juliet~M. Corbin} {and} \bibinfo{person}{Anselm Strauss}.} \bibinfo{year}{1990}\natexlab{}.
\newblock \showarticletitle{Grounded {Theory} {Research}: {Procedures}, {Canons}, and {Evaluative} {Criteria}}.
\newblock \bibinfo{journal}{\emph{Qualitative Sociology}} \bibinfo{volume}{13}, \bibinfo{number}{1} (\bibinfo{date}{March} \bibinfo{year}{1990}).
\newblock
\urldef\tempurl%
\url{https://doi.org/10.1007/BF00988593}
\showDOI{\tempurl}


\bibitem[Creswell and Clark(2007)]%
        {creswell_designing_2007}
\bibfield{author}{\bibinfo{person}{John~W. Creswell} {and} \bibinfo{person}{Vicki L.~Plano Clark}.} \bibinfo{year}{2007}\natexlab{}.
\newblock \bibinfo{booktitle}{\emph{Designing and {Conducting} {Mixed} {Methods} {Research}}}.
\newblock \bibinfo{publisher}{SAGE Publications}, \bibinfo{address}{Thousand Oaks, US}.
\newblock


\bibitem[Dupree et~al\mbox{.}(2016)]%
        {dupree_privacy_2016}
\bibfield{author}{\bibinfo{person}{Janna~L. Dupree}, \bibinfo{person}{Richard Devries}, \bibinfo{person}{Daniel~M. Berry}, {and} \bibinfo{person}{Edward Lank}.} \bibinfo{year}{2016}\natexlab{}.
\newblock \showarticletitle{Privacy {Personas}: {Clustering} {Users} via {Attitudes} and {Behaviors} toward {Security} {Practices}}. In \bibinfo{booktitle}{\emph{Proceedings of the {CHI} {Conference} on {Human} {Factors} in {Computing} {Systems}}} \emph{(\bibinfo{series}{{CHI} 2016})}. \bibinfo{publisher}{ACM}, \bibinfo{address}{San Jose, US}, \bibinfo{pages}{5228--5239}.
\newblock
\urldef\tempurl%
\url{https://doi.org/10.1145/2858036.2858214}
\showDOI{\tempurl}


\bibitem[Elueze and Quan-Haase(2018)]%
        {elueze_privacy_2018}
\bibfield{author}{\bibinfo{person}{Isioma Elueze} {and} \bibinfo{person}{Anabel Quan-Haase}.} \bibinfo{year}{2018}\natexlab{}.
\newblock \showarticletitle{Privacy {Attitudes} and {Concerns} in the {Digital} {Lives} of {Older} {Adults}: {Westin}’s {Privacy} {Attitude} {Typology} {Revisited}}.
\newblock \bibinfo{journal}{\emph{American Behavioral Scientist}} \bibinfo{volume}{62}, \bibinfo{number}{10} (\bibinfo{date}{Sept.} \bibinfo{year}{2018}), \bibinfo{pages}{1372--1391}.
\newblock
\urldef\tempurl%
\url{https://doi.org/10.1177/0002764218787026}
\showDOI{\tempurl}


\bibitem[Ferrarello et~al\mbox{.}(2022)]%
        {ferrarello_reframing_2022}
\bibfield{author}{\bibinfo{person}{Laura Ferrarello}, \bibinfo{person}{Rute Fiadeiro}, \bibinfo{person}{Riccardo Mazzon}, {and} \bibinfo{person}{Andrea Cavallaro}.} \bibinfo{year}{2022}\natexlab{}.
\newblock \showarticletitle{Reframing the {Narrative} of {Privacy} {Through} {System}-{Thinking} {Design}}. In \bibinfo{booktitle}{\emph{Proceedings of the Design Research Society}} \emph{(\bibinfo{series}{{DRS} 2022})}. \bibinfo{publisher}{Design Research Society}, \bibinfo{address}{Bilbao, SPN}.
\newblock
\urldef\tempurl%
\url{https://doi.org/10.21606/drs.2022.620}
\showDOI{\tempurl}


\bibitem[Fowlkes and Mallows(1983)]%
        {fowlkes_method_1983}
\bibfield{author}{\bibinfo{person}{Edward~B. Fowlkes} {and} \bibinfo{person}{Colin~L. Mallows}.} \bibinfo{year}{1983}\natexlab{}.
\newblock \showarticletitle{A {Method} for {Comparing} {Two} {Hierarchical} {Clusterings}}.
\newblock \bibinfo{journal}{\emph{J. Amer. Statist. Assoc.}} \bibinfo{volume}{78}, \bibinfo{number}{383} (\bibinfo{date}{Sept.} \bibinfo{year}{1983}), \bibinfo{pages}{553--569}.
\newblock
\urldef\tempurl%
\url{https://doi.org/10.1080/01621459.1983.10478008}
\showDOI{\tempurl}


\bibitem[Gallardo et~al\mbox{.}(2023)]%
        {gallardo_speculative_2023}
\bibfield{author}{\bibinfo{person}{Andrea Gallardo}, \bibinfo{person}{Chris Choy}, \bibinfo{person}{Jaideep Juneja}, \bibinfo{person}{Efe Bozkir}, \bibinfo{person}{Camille Cobb}, \bibinfo{person}{Lujo Bauer}, {and} \bibinfo{person}{Lorrie~F. Cranor}.} \bibinfo{year}{2023}\natexlab{}.
\newblock \showarticletitle{Speculative {Privacy} {Concerns} {About} {AR} {Glasses} {Data} {Collection}}.
\newblock \bibinfo{journal}{\emph{Proceedings on Privacy Enhancing Technologies}} \bibinfo{number}{4} (\bibinfo{date}{Oct.} \bibinfo{year}{2023}), \bibinfo{pages}{416--435}.
\newblock
\urldef\tempurl%
\url{https://doi.org/10.56553/popets-2023-0117}
\showDOI{\tempurl}


\bibitem[Gibbs(2018)]%
        {gibbs_analyzing_2018}
\bibfield{author}{\bibinfo{person}{Graham~R. Gibbs}.} \bibinfo{year}{2018}\natexlab{}.
\newblock \bibinfo{booktitle}{\emph{Analyzing {Qualitative} {Data}}}.
\newblock \bibinfo{publisher}{SAGE Publications}, \bibinfo{address}{London, UK}.
\newblock
\urldef\tempurl%
\url{https://doi.org/10.4135/9781526441867}
\showDOI{\tempurl}


\bibitem[Goyeneche et~al\mbox{.}(2024)]%
        {goyeneche_linked_2024}
\bibfield{author}{\bibinfo{person}{David Goyeneche}, \bibinfo{person}{Stephen Singaraju}, {and} \bibinfo{person}{Luis Arango}.} \bibinfo{year}{2024}\natexlab{}.
\newblock \showarticletitle{Linked by {Age}: {A} {Study} on {Social} {Media} {Privacy} {Concerns} {Among} {Younger} and {Older} {Adults}}.
\newblock \bibinfo{journal}{\emph{Industrial Management \& Data Systems}} \bibinfo{volume}{124}, \bibinfo{number}{2} (\bibinfo{date}{Jan.} \bibinfo{year}{2024}), \bibinfo{pages}{640--665}.
\newblock
\urldef\tempurl%
\url{https://doi.org/10.1108/IMDS-07-2023-0462}
\showDOI{\tempurl}


\bibitem[Hennink et~al\mbox{.}(2017)]%
        {hennink_code_2017}
\bibfield{author}{\bibinfo{person}{Monique~M. Hennink}, \bibinfo{person}{Bonnie~N. Kaiser}, {and} \bibinfo{person}{Vincent~C. Marconi}.} \bibinfo{year}{2017}\natexlab{}.
\newblock \showarticletitle{Code {Saturation} {Versus} {Meaning} {Saturation}: {How} {Many} {Interviews} {Are} {Enough}?}
\newblock \bibinfo{journal}{\emph{Qualitative Health Research}} \bibinfo{volume}{27}, \bibinfo{number}{4} (\bibinfo{date}{March} \bibinfo{year}{2017}), \bibinfo{pages}{591--608}.
\newblock
\urldef\tempurl%
\url{https://doi.org/10.1177/1049732316665344}
\showDOI{\tempurl}


\bibitem[Jansen et~al\mbox{.}(2020)]%
        {jansen_data-driven_2020}
\bibfield{author}{\bibinfo{person}{Bernard~J. Jansen}, \bibinfo{person}{Joni~O. Salminen}, {and} \bibinfo{person}{Soon-Gyo Jung}.} \bibinfo{year}{2020}\natexlab{}.
\newblock \showarticletitle{Data-{Driven} {Personas} for {Enhanced} {User} {Understanding}: {Combining} {Empathy} with {Rationality} for {Better} {Insights} to {Analytics}}.
\newblock \bibinfo{journal}{\emph{Data and Information Management}} \bibinfo{volume}{4}, \bibinfo{number}{1} (\bibinfo{date}{March} \bibinfo{year}{2020}).
\newblock
\urldef\tempurl%
\url{https://doi.org/10.2478/dim-2020-0005}
\showDOI{\tempurl}


\bibitem[Kaufman and Rousseeuw(2005)]%
        {kaufman_finding_2005}
\bibfield{author}{\bibinfo{person}{Leonard Kaufman} {and} \bibinfo{person}{Peter~J. Rousseeuw}.} \bibinfo{year}{2005}\natexlab{}.
\newblock \bibinfo{booktitle}{\emph{Finding {Groups} in {Data}: an {Introduction} to {Cluster} {Analysis}}}.
\newblock \bibinfo{publisher}{Wiley}, \bibinfo{address}{Hoboken, US}.
\newblock


\bibitem[King(2014)]%
        {king_taken_2014}
\bibfield{author}{\bibinfo{person}{Jennifer King}.} \bibinfo{year}{2014}\natexlab{}.
\newblock \showarticletitle{Taken {Out} of {Context}: an {Empirical} {Analysis} of {Westin}’s {Privacy} {Scale}}. In \bibinfo{booktitle}{\emph{Proceedings of the Workshop on {Privacy} {Personas} and {Segmentation}}} \emph{(\bibinfo{series}{{SOUPS} 2014})}. \bibinfo{publisher}{USENIX Association}, \bibinfo{address}{Menlo Park, US}.
\newblock
\urldef\tempurl%
\url{https://api.semanticscholar.org/CorpusID:481189}
\showURL{%
\tempurl}


\bibitem[Li et~al\mbox{.}(2020)]%
        {li_towards_2020}
\bibfield{author}{\bibinfo{person}{Yifang Li}, \bibinfo{person}{Nishant Vishwamitra}, \bibinfo{person}{Hongxin Hu}, {and} \bibinfo{person}{Kelly Caine}.} \bibinfo{year}{2020}\natexlab{}.
\newblock \showarticletitle{Towards a {Taxonomy} of {Content} {Sensitivity} and {Sharing} {Preferences} for {Photos}}. In \bibinfo{booktitle}{\emph{Proceedings of the {CHI} {Conference} on {Human} {Factors} in {Computing} {Systems}}}. \bibinfo{publisher}{ACM}, \bibinfo{address}{Honolulu, US}.
\newblock
\urldef\tempurl%
\url{https://doi.org/10.1145/3313831.3376498}
\showDOI{\tempurl}


\bibitem[Lin et~al\mbox{.}(2014)]%
        {lin_modeling_2014}
\bibfield{author}{\bibinfo{person}{Jialiu Lin}, \bibinfo{person}{Bin Liu}, \bibinfo{person}{Norman Sadeh}, {and} \bibinfo{person}{Jason~I. Hong}.} \bibinfo{year}{2014}\natexlab{}.
\newblock \showarticletitle{Modeling {Users}’ Mobile App Privacy Preferences: {Restoring} Usability in a Sea of Permission Settings}. In \bibinfo{booktitle}{\emph{Proceedings of the {Tenth} {USENIX} {Conference} on {Usable} {Privacy} and {Security}}} \emph{(\bibinfo{series}{{SOUPS} 2014})}. \bibinfo{publisher}{USENIX Association}, \bibinfo{address}{Menlo Park, US}, \bibinfo{pages}{199--212}.
\newblock


\bibitem[Lucero(2015)]%
        {lucero_using_2015}
\bibfield{author}{\bibinfo{person}{Andrés Lucero}.} \bibinfo{year}{2015}\natexlab{}.
\newblock \showarticletitle{Using {Affinity} {Diagrams} to {Evaluate} {Interactive} {Prototypes}}. In \bibinfo{booktitle}{\emph{Proceedings of the Human-{Computer} {Interaction}}} \emph{(\bibinfo{series}{{INTERACT} 2015})}. \bibinfo{publisher}{Springer International Publishing}, \bibinfo{address}{Bamberg, GER}, \bibinfo{pages}{231--248}.
\newblock
\urldef\tempurl%
\url{https://doi.org/10.1007/978-3-319-22668-2_19}
\showDOI{\tempurl}


\bibitem[Milne et~al\mbox{.}(2017)]%
        {milne_information_2017}
\bibfield{author}{\bibinfo{person}{George~R. Milne}, \bibinfo{person}{George Pettinico}, \bibinfo{person}{Fatima Hajjat}, {and} \bibinfo{person}{Ereni Markos}.} \bibinfo{year}{2017}\natexlab{}.
\newblock \showarticletitle{Information {Sensitivity} {Typology}: {Mapping} the {Degree} and {Type} of {Risk} {Consumers} {Perceive} in {Personal} {Data} {Sharing}}.
\newblock \bibinfo{journal}{\emph{Journal of Consumer Affairs}} \bibinfo{volume}{51}, \bibinfo{number}{1} (\bibinfo{date}{March} \bibinfo{year}{2017}), \bibinfo{pages}{133--161}.
\newblock
\urldef\tempurl%
\url{https://doi.org/10.1111/joca.12111}
\showDOI{\tempurl}


\bibitem[Mugan et~al\mbox{.}(2011)]%
        {mugan_understandable_2011}
\bibfield{author}{\bibinfo{person}{Jonathan Mugan}, \bibinfo{person}{Tarun Sharma}, {and} \bibinfo{person}{Norman Sadeh}.} \bibinfo{year}{2011}\natexlab{}.
\newblock \bibinfo{booktitle}{\emph{Understandable Learning of Privacy Preferences Through Default Personas and Suggestions}}.
\newblock \bibinfo{type}{{T}echnical {R}eport}. \bibinfo{address}{Pittsburgh, US}.
\newblock


\bibitem[Novick and Lewis(1967)]%
        {novick_coefficient_1967}
\bibfield{author}{\bibinfo{person}{Melvin~R. Novick} {and} \bibinfo{person}{Charles Lewis}.} \bibinfo{year}{1967}\natexlab{}.
\newblock \showarticletitle{Coefficient Alpha and the Reliability of Composite Measurements}.
\newblock \bibinfo{journal}{\emph{Psychometrika}} \bibinfo{volume}{32}, \bibinfo{number}{1} (\bibinfo{date}{March} \bibinfo{year}{1967}).
\newblock
\urldef\tempurl%
\url{https://doi.org/10.1007/BF02289400}
\showDOI{\tempurl}


\bibitem[{Ponnurangam Kumaraguru} and Cranor(2005)]%
        {ponnurangam__kumaraguru_privacy_2005}
\bibfield{author}{\bibinfo{person}{{Ponnurangam Kumaraguru}} {and} \bibinfo{person}{Lorrie~F. Cranor}.} \bibinfo{year}{2005}\natexlab{}.
\newblock \bibinfo{booktitle}{\emph{Privacy indexes: a survey of {Westin}'s studies}}.
\newblock \bibinfo{type}{{ISRI}} CMU-ISRI-05-138. \bibinfo{institution}{Institute for Software Research International, Carnegie Mellon University}, \bibinfo{address}{Pittsburgh, US}.
\newblock
\urldef\tempurl%
\url{http://reports-archive.adm.cs.cmu.edu/anon/anon/home/ftp/usr0/ftp/isri2005/CMU-ISRI-05-138.pdf}
\showURL{%
\tempurl}


\bibitem[{Prolific}(2024)]%
        {prolific_prolific_2024}
\bibfield{author}{\bibinfo{person}{{Prolific}}.} \bibinfo{year}{2024}\natexlab{}.
\newblock \bibinfo{booktitle}{\emph{Prolific}}.
\newblock London, UK.
\newblock
\urldef\tempurl%
\url{https://www.prolific.co}
\showURL{%
Retrieved Mar. 1, 2024 from \tempurl}


\bibitem[Schomakers et~al\mbox{.}(2019)]%
        {schomakers_typology_2019}
\bibfield{author}{\bibinfo{person}{Eva-Maria Schomakers}, \bibinfo{person}{Chantal Lidynia}, {and} \bibinfo{person}{Martina Ziefle}.} \bibinfo{year}{2019}\natexlab{}.
\newblock \showarticletitle{A {Typology} of {Online} {Privacy} {Personalities}: {Exploring} and {Segmenting} {Users}’ {Diverse} {Privacy} {Attitudes} and {Behaviors}}.
\newblock \bibinfo{journal}{\emph{Journal of Grid Computing}} \bibinfo{volume}{17}, \bibinfo{number}{4} (\bibinfo{date}{Dec.} \bibinfo{year}{2019}), \bibinfo{pages}{727--747}.
\newblock
\urldef\tempurl%
\url{https://doi.org/10.1007/s10723-019-09500-3}
\showDOI{\tempurl}


\bibitem[Stöver et~al\mbox{.}(2023a)]%
        {stover_how_2023}
\bibfield{author}{\bibinfo{person}{Alina Stöver}, \bibinfo{person}{Nina Gerber}, \bibinfo{person}{Henning Pridöhl}, \bibinfo{person}{Max Maass}, \bibinfo{person}{Sebastian Bretthauer}, \bibinfo{person}{Indra Spiecker~gen. Döhmann}, \bibinfo{person}{Matthias Hollick}, {and} \bibinfo{person}{Dominik Herrmann}.} \bibinfo{year}{2023}\natexlab{a}.
\newblock \showarticletitle{How {Website} {Owners} {Face} {Privacy} {Issues}: {Thematic} {Analysis} of {Responses} from a {Covert} {Notification} {Study} {Reveals} {Diverse} {Circumstances} and {Challenges}}.
\newblock \bibinfo{journal}{\emph{Proceedings on Privacy Enhancing Technologies}} \bibinfo{number}{2} (\bibinfo{date}{April} \bibinfo{year}{2023}), \bibinfo{pages}{251--264}.
\newblock
\urldef\tempurl%
\url{https://doi.org/10.56553/popets-2023-0051}
\showDOI{\tempurl}


\bibitem[Stöver et~al\mbox{.}(2023b)]%
        {stover_investigating_2023}
\bibfield{author}{\bibinfo{person}{Alina Stöver}, \bibinfo{person}{Sara Hahn}, \bibinfo{person}{Felix Kretschmer}, {and} \bibinfo{person}{Nina Gerber}.} \bibinfo{year}{2023}\natexlab{b}.
\newblock \showarticletitle{Investigating how {Users} {Imagine} their {Personal} {Privacy} {Assistant}}.
\newblock \bibinfo{journal}{\emph{Proceedings on Privacy Enhancing Technologies}} \bibinfo{number}{2} (\bibinfo{date}{April} \bibinfo{year}{2023}), \bibinfo{pages}{384--402}.
\newblock
\urldef\tempurl%
\url{https://doi.org/10.56553/popets-2023-0059}
\showDOI{\tempurl}


\bibitem[Tukey(1977)]%
        {tukey_exploratory_1977}
\bibfield{author}{\bibinfo{person}{John~W. Tukey}.} \bibinfo{year}{1977}\natexlab{}.
\newblock \bibinfo{booktitle}{\emph{Exploratory {Data} {Analysis}}}.
\newblock \bibinfo{publisher}{Addison-Wesley}, \bibinfo{address}{Reading, US; London, UK}.
\newblock


\bibitem[Urban(2014)]%
        {urban_privacy_2014}
\bibfield{author}{\bibinfo{person}{Jennifer Urban}.} \bibinfo{year}{2014}\natexlab{}.
\newblock \showarticletitle{The {Privacy} {Pragmatic} as {Privacy} {Vulnerable}}. In \bibinfo{booktitle}{\emph{Proceedings of the Workshop on {Privacy} {Personas} and {Segmentation}}} \emph{(\bibinfo{series}{{SOUPS} 2014})}. \bibinfo{address}{Menlo Park, US.}
\newblock
\urldef\tempurl%
\url{https://doi.org/10.31235/osf.io/yh8nj}
\showDOI{\tempurl}


\bibitem[{U.S. Government Publishing Office}(2001)]%
        {us_government_publishing_office_opinion_2001}
\bibfield{author}{\bibinfo{person}{{U.S. Government Publishing Office}}.} \bibinfo{year}{2001}\natexlab{}.
\newblock \bibinfo{booktitle}{\emph{Opinion {Surveys}: {What} {Consumers} {Have} to {Say} {About} {Information} {Privacy}}}.
\newblock
\urldef\tempurl%
\url{https://www.govinfo.gov/content/pkg/CHRG-107hhrg72825/html/CHRG-107hhrg72825.htm}
\showURL{%
Retrieved Mar. 1, 2024 from \tempurl}


\bibitem[Wisniewski et~al\mbox{.}(2014)]%
        {wisniewski_profiling_2014}
\bibfield{author}{\bibinfo{person}{Pamela Wisniewski}, \bibinfo{person}{Bart~P. Knijnenburg}, {and} \bibinfo{person}{Heather~R. Lipford}.} \bibinfo{year}{2014}\natexlab{}.
\newblock \showarticletitle{Profiling {Facebook} {Users} {Privacy} {Behaviors}}. In \bibinfo{booktitle}{\emph{Proceedings of the Workshop on {Privacy} {Personas} and {Segmentation}}} \emph{(\bibinfo{series}{{SOUPS} 2014})}. \bibinfo{publisher}{USENIX Association}, \bibinfo{address}{Menlo Park, US.}
\newblock


\bibitem[Xu et~al\mbox{.}(2023)]%
        {xu_dipa2_2023}
\bibfield{author}{\bibinfo{person}{Anran Xu}, \bibinfo{person}{Zhongyi Zhou}, \bibinfo{person}{Kakeru Miyazaki}, \bibinfo{person}{Ryo Yoshikawa}, \bibinfo{person}{Simo Hosio}, {and} \bibinfo{person}{Koji Yatani}.} \bibinfo{year}{2023}\natexlab{}.
\newblock \showarticletitle{{DIPA2}: {An} {Image} {Dataset} with {Cross}-{Cultural} {Privacy} {Perception} {Annotations}}.
\newblock \bibinfo{journal}{\emph{Proceedings of the ACM on Interactive, Mobile, Wearable and Ubiquitous Technologies}} \bibinfo{volume}{7}, \bibinfo{number}{4} (\bibinfo{date}{Dec.} \bibinfo{year}{2023}).
\newblock
\urldef\tempurl%
\url{https://doi.org/10.1145/3631439}
\showDOI{\tempurl}


\bibitem[Xu and Lee(2020)]%
        {xu_identifying_2020}
\bibfield{author}{\bibinfo{person}{Yu Xu} {and} \bibinfo{person}{Michael~J. Lee}.} \bibinfo{year}{2020}\natexlab{}.
\newblock \showarticletitle{Identifying {Personas} in {Online} {Shopping} {Communities}}.
\newblock \bibinfo{journal}{\emph{Multimodal Technologies and Interaction}} \bibinfo{volume}{4}, \bibinfo{number}{2} (\bibinfo{date}{May} \bibinfo{year}{2020}).
\newblock
\urldef\tempurl%
\url{https://doi.org/10.3390/mti4020019}
\showDOI{\tempurl}


\bibitem[Zeissig et~al\mbox{.}(2017)]%
        {zhou_online_2017}
\bibfield{author}{\bibinfo{person}{Eva-Maria Zeissig}, \bibinfo{person}{Chantal Lidynia}, \bibinfo{person}{Luisa Vervier}, \bibinfo{person}{Andera Gadeib}, {and} \bibinfo{person}{Martina Ziefle}.} \bibinfo{year}{2017}\natexlab{}.
\newblock \showarticletitle{Online {Privacy} {Perceptions} of {Older} {Adults}}. In \bibinfo{booktitle}{\emph{Proceedings of the Human {Aspects} of {IT} for the {Aged} {Population}}} \emph{(\bibinfo{series}{{ITAP} 2017})}. \bibinfo{publisher}{Springer International Publishing}, \bibinfo{address}{Vancouver, CAN}, \bibinfo{pages}{181--200}.
\newblock
\urldef\tempurl%
\url{https://doi.org/10.1007/978-3-319-58536-9_16}
\showDOI{\tempurl}


\bibitem[Zerr et~al\mbox{.}(2012)]%
        {zerr_picalert_2012}
\bibfield{author}{\bibinfo{person}{Sergej Zerr}, \bibinfo{person}{Stefan Siersdorfer}, {and} \bibinfo{person}{Jonathon Hare}.} \bibinfo{year}{2012}\natexlab{}.
\newblock \showarticletitle{{PicAlert}!: a {System} for {Privacy}-{Aware} {Image} {Classification} and {Retrieval}}. In \bibinfo{booktitle}{\emph{Proceedings of the Twenty-First {ACM} international conference on {Information} and knowledge management}}. \bibinfo{publisher}{ACM}, \bibinfo{address}{Maui, US}, \bibinfo{pages}{2710--2712}.
\newblock
\showISBNx{978-1-4503-1156-4}
\urldef\tempurl%
\url{https://doi.org/10.1145/2396761.2398735}
\showDOI{\tempurl}


\bibitem[Zhao et~al\mbox{.}(2022)]%
        {zhao_privacyalert_2022}
\bibfield{author}{\bibinfo{person}{Chenye Zhao}, \bibinfo{person}{Jasmine Mangat}, \bibinfo{person}{Sujay Koujalgi}, \bibinfo{person}{Anna Squicciarini}, {and} \bibinfo{person}{Cornelia Caragea}.} \bibinfo{year}{2022}\natexlab{}.
\newblock \showarticletitle{{PrivacyAlert}: a {Dataset} for {Image} {Privacy} {Prediction}}. In \bibinfo{booktitle}{\emph{Proceedings of the International AAAI Conference on Web and Social Media}}, Vol.~\bibinfo{volume}{16}. \bibinfo{publisher}{AAAI Press}, \bibinfo{address}{Washington, US}, \bibinfo{pages}{1352--1361}.
\newblock
\showISSN{2334-0770, 2162-3449}
\urldef\tempurl%
\url{https://doi.org/10.1609/icwsm.v16i1.19387}
\showDOI{\tempurl}


\end{thebibliography}

\onecolumn
\appendix
\section*{Appendices}
\section{Sensitivity Analysis}
\label{app:sensitivity_analysis}
\begin{figure*}[h]
    \centering
    \includegraphics[width = 0.45\linewidth]{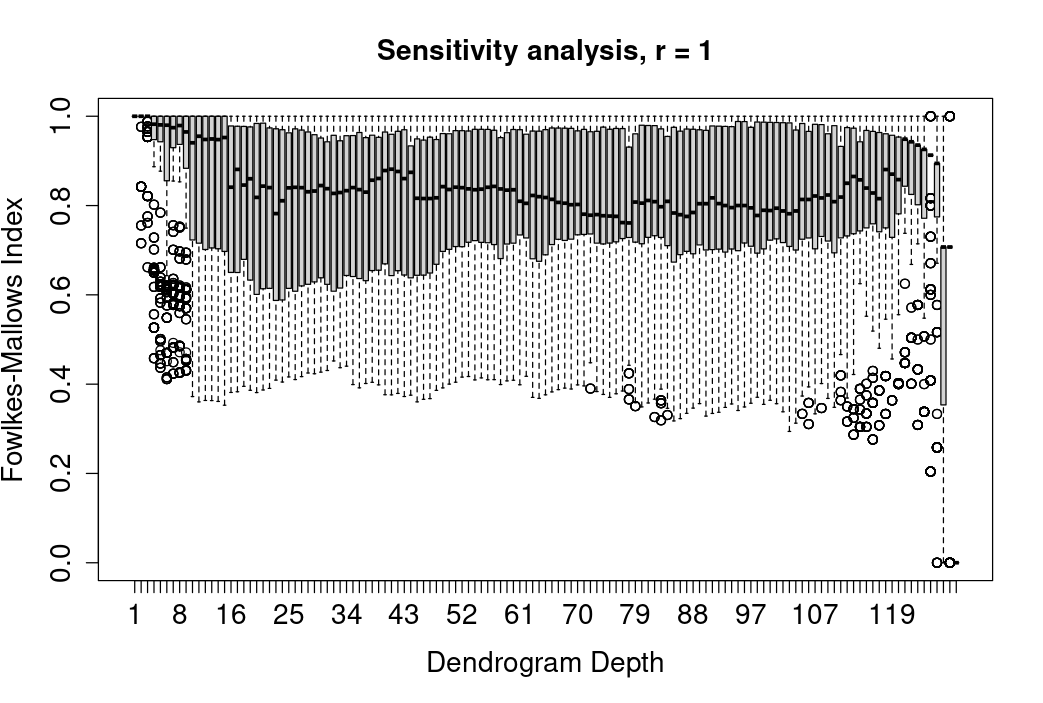}
    \includegraphics[width = 0.45\linewidth]{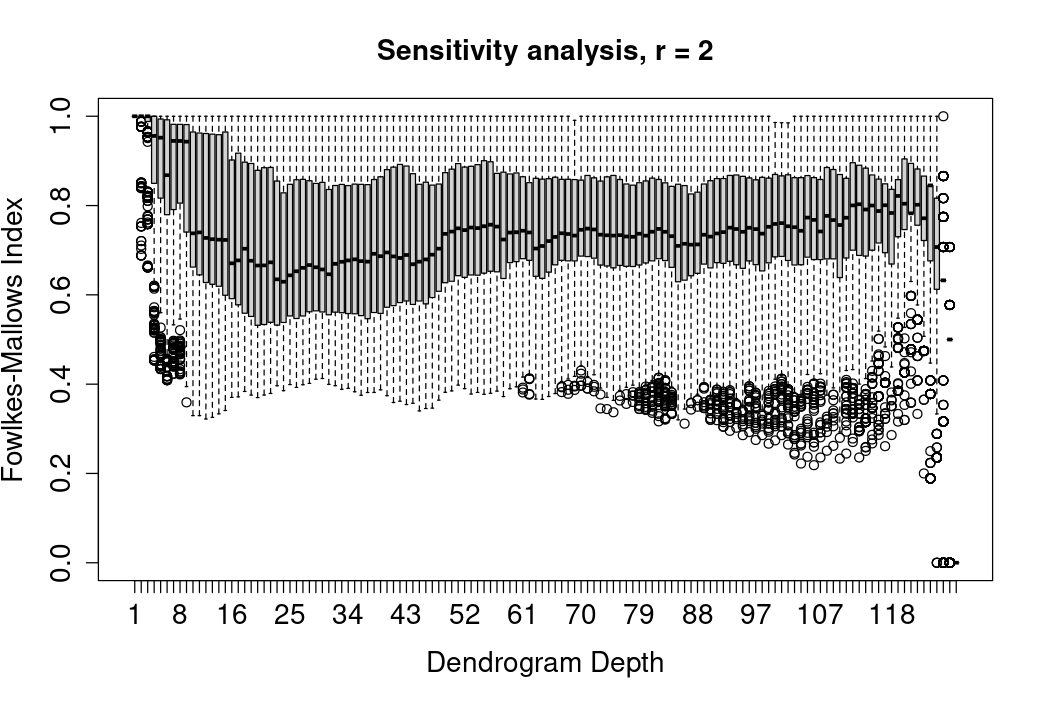}
    \includegraphics[width = 0.45\linewidth]{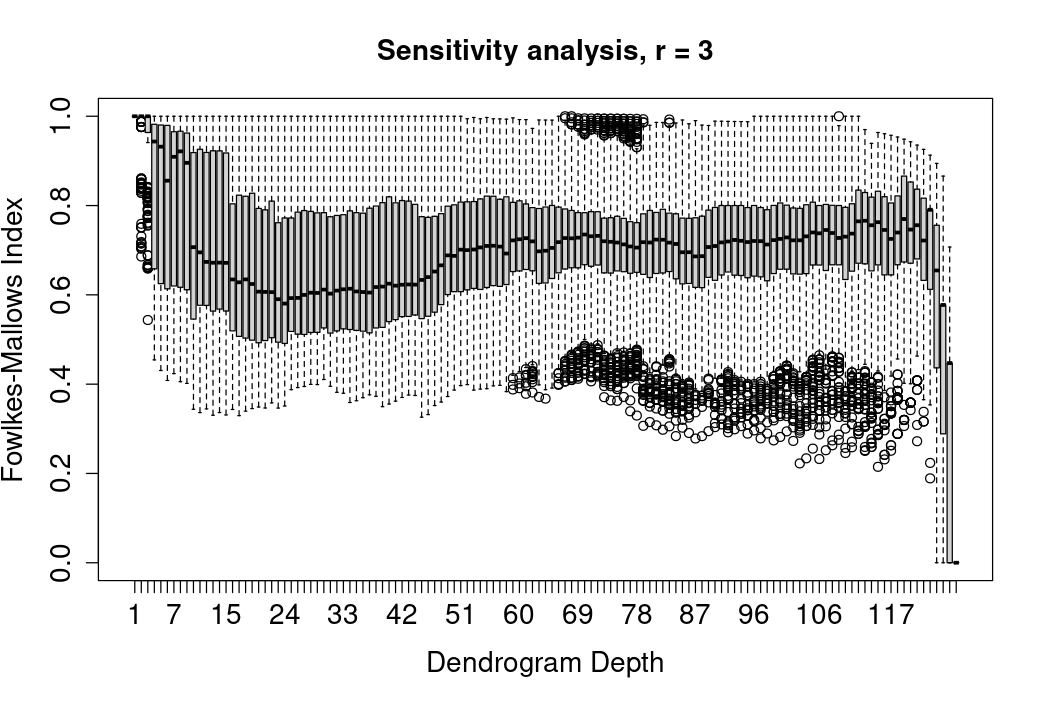}
    \includegraphics[width = 0.45\linewidth]{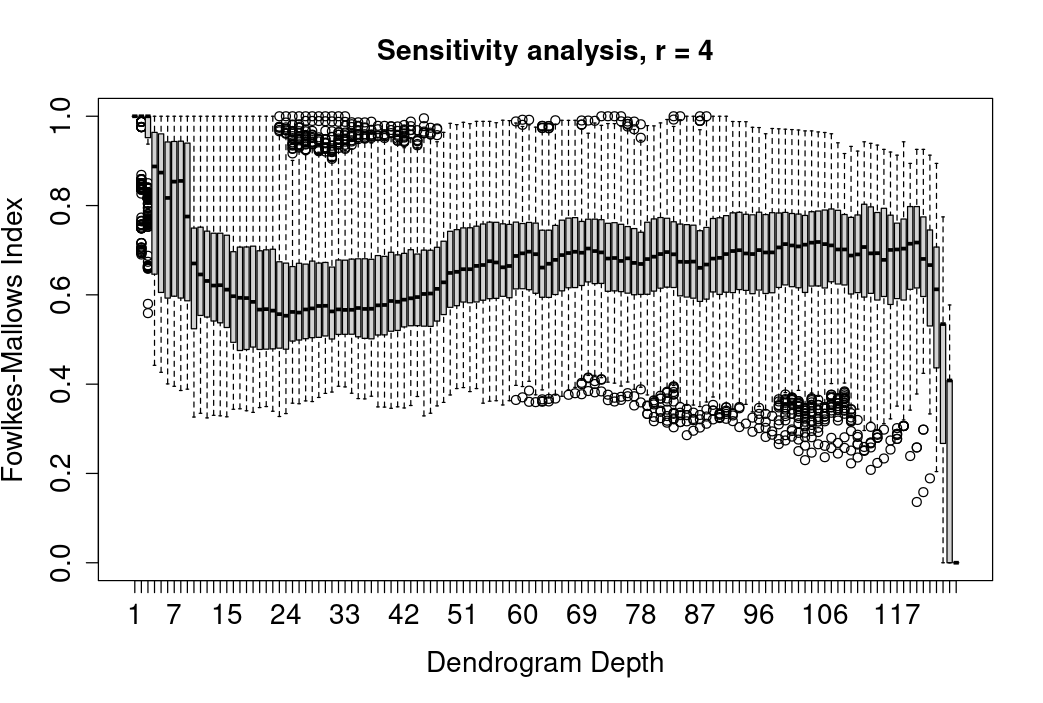}
    \includegraphics[width = 0.45\linewidth]{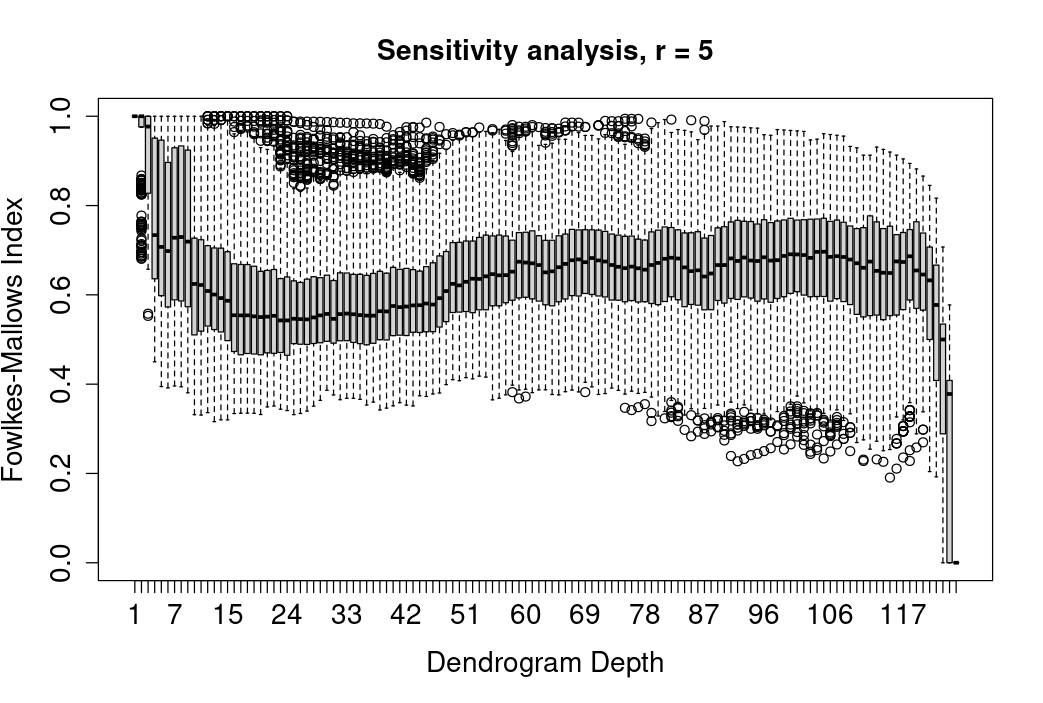}
    \includegraphics[width = 0.45\linewidth]{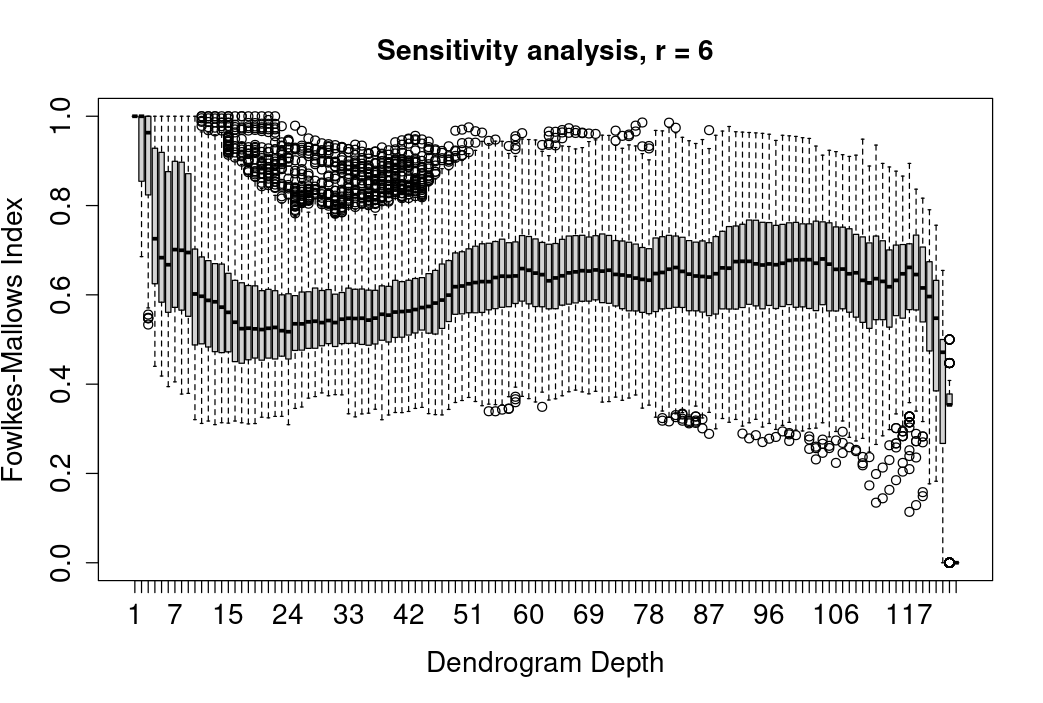}
    \caption{Extended sensitivity analysis of the dendrogram obtained on the generation set $\mathscr{G}$.  We randomly remove $r$ participants from the generation set $\mathscr{G}$, form a new dendrogram, and compute the Fowlkes-Mallows (FM) Index~\citep{fowlkes_method_1983} between the newly obtained dendrogram and a dendrogram obtained on the initial generation set $\mathscr{G}$. We repeat the sampling procedure 500 times for each value of $r$, and report the distribution of the FM Index.
    }
    \label{fig:FM_dist}
\end{figure*}

\newpage
\twocolumn
\section{Explanatory variables}
\label{app:explanatory_vars}
\begin{enumerate}[label=$b_{\arabic*}$, font=\footnotesize, before=\footnotesize, after=\footnotesize]
  
  \item[$l_1$] \textbf{sharing}
      \begin{enumerate}[label=Trait \arabic*, font=\footnotesize, before=\footnotesize, after=\footnotesize]
            \item[$t_{1,2,3}$]{has shown to share a few / moderate / a lot pieces of personal information in the last 10 images}
      \end{enumerate}

  \item[$l_2$] \textbf{sensitivity}
      \begin{enumerate}[label=Trait \arabic*, font=\footnotesize, before=\footnotesize, after=\footnotesize]
        \item[$t_{4\mbox{-}8}$]{has a low / low medium / medium / medium high / high level of sensitivity regarding the private information}  
      \end{enumerate}

  \item[$l_3$] \textbf{privacy protection importance (initial)}
      \begin{enumerate}[label=Trait \arabic*, font=\footnotesize, before=\footnotesize, after=\footnotesize]
        \item[$t_{9\mbox{-}13}$]{initially believes that privacy protection is unimportant / not that important / no opinion / moderately important / very important}  
      \end{enumerate}

  \item[$l_4$] \textbf{ how easy it is to protect}
      \begin{enumerate}[label=Trait \arabic*, font=\footnotesize, before=\footnotesize, after=\footnotesize]
        \item[$t_{14\mbox{-}18}$]{initially thinks that protecting personal information is extremely easy / mostly easy / moderately easy / hard / very hard with the tools that are available to them}  
      \end{enumerate}

  \item[$l_5$]\textbf{desired level of control}
      \begin{enumerate}[label=Trait \arabic*, font=\footnotesize, before=\footnotesize, after=\footnotesize]
        \item[$t_{19\mbox{-}23}$]{initially expressed that they never / rarely / sometimes / often / always want to be in control of their personal information}  
      \end{enumerate}
      
  \item[$l_6$]\textbf{knowledge about checking for sensitive content}
      \begin{enumerate}[label=Trait \arabic*, font=\footnotesize, before=\footnotesize, after=\footnotesize]
        \item[$t_{24, 25 ,26}$]{illustrated no awareness / only awareness / familiarity of social media checking images for sensitive content}  
      \end{enumerate}

  \item[$l_7$] \textbf{knowledge about awareness for personal information extraction}
      \begin{enumerate}[label=Trait \arabic*, font=\footnotesize, before=\footnotesize, after=\footnotesize]
        \item[$t_{27, 28 ,29}$]{illustrated no awareness / only awareness / familiarity of social media being able to extract personal information from an image and generate a personal profile on a user}  
      \end{enumerate}

  \item[$l_8$]\textbf{knowledge about what is happening with the data}
      \begin{enumerate}[label=Trait \arabic*, font=\footnotesize, before=\footnotesize, after=\footnotesize]
        \item[$t_{30\mbox{-}34}$]{claimed to have no understanding / little understanding / moderate understanding / to mostly understand / full understanding of what is happening to their images when uploading them online}  
      \end{enumerate}

  \item[$l_9$] \textbf{perceived level of control}
      \begin{enumerate}[label=Trait \arabic*, font=\footnotesize, before=\footnotesize, after=\footnotesize]
        \item[$t_{35\mbox{-}39}$]{after being exposed to how companies can extract data from their images, they feel that they have no control / only little control / moderate control / can mostly control / full control over their personal information when sharing images online}  
      \end{enumerate}

  \item[$l_{10}$] \textbf{usefulness of the PET}
      \begin{enumerate}[label=Trait \arabic*, font=\footnotesize, before=\footnotesize, after=\footnotesize]
        \item[$t_{40, 41, 42}$]{they feel that the proposed privacy enhancing technology will not help them / not sure will help them / will help them to protect their privacy when sharing images online}  
      \end{enumerate}

  \item[$l_{11}$] \textbf{would use the PET}
      \begin{enumerate}[label=Trait \arabic*, font=\footnotesize, before=\footnotesize, after=\footnotesize]
        \item[$t_{43\mbox{-}47}$]{they would never / rarely / sometimes / often / always use the proposed privacy enhancing filter when sharing images online}  
      \end{enumerate}

  \item[$l_{12}$] \textbf{privacy protection importance (end)}
      \begin{enumerate}[label=Trait \arabic*, font=\footnotesize, before=\footnotesize, after=\footnotesize]
        \item[$t_{48\mbox{-}52}$]{after being exposed to companies being able to extract personal information and a proposed privacy enhancing tool, they believe that privacy protection is unimportant / not that important / no opinion / moderately important / very important}  
      \end{enumerate}

  \item[$l_{13}$] \textbf{privacy protection importance (delta)}
      \begin{enumerate}[label=Trait \arabic*, font=\footnotesize, before=\footnotesize, after=\footnotesize]
        \item[$t_{53\mbox{-}59}$]{drastic decrease / significant decrease / slight increase / no change / slight decrease / significant increase / drastic increase in privacy protection importance after being exposed to companies being able to extract personal information and a proposed privacy enhancing tool}  
      \end{enumerate}

  \item[$l_{14}$] \textbf{control mismatch}
      \begin{enumerate}[label=Trait \arabic*, font=\footnotesize, before=\footnotesize, after=\footnotesize]
        \item[$t_{60\mbox{-}66}$]{extreme (less control than wanted) / significant (less control than wanted) / slight (less control than wanted) / no / slight (more control than wanted) / significant (more control than wanted) / extreme (more control than wanted) mismatch between the perceived and desired level of control over personal information after being exposed to companies being able to extract personal information}  
      \end{enumerate}

\item{$t_{67}$: demonstrated a high level of knowledge in some aspects}
\item{$t_{68}$: demonstrated a lack of knowledge in some aspects}
\item{$t_{69}$: appreciates that before experience, did not realise all the potential risks when sharing their image online}
\item{$t_{70}$: after being exposed to how companies can extract data from their images, the participant felt surprised}
\item{$t_{71}$: after being exposed to how companies can extract data from their images, the participant felt confused / lost}
\item{$t_{72}$: after being exposed to how companies can extract data from their images, the participant felt unimpressed (extracted information was generic)}
\item{$t_{73}$: after being exposed to how companies can extract data from their images, the participant felt uncomfortable / scared}
\item{$t_{74}$: after being exposed to how companies can extract data from their images, the participant felt angry / violated / offended}
\item{$t_{75}$: after being exposed to how companies can extract data from their images, the participant did not feel surprised (was aware of this happening)}
\item{$t_{76}$: has expressed that physical security is an element of private information for them}
\item{$t_{77}$: has expressed that financial status is an element of private information for them}
\item{$t_{78}$: has expressed concern about protecting the loved ones/friends}
\item{$t_{79}$: has expressed that sexual orientation is an element of private information for them}
\item{$t_{80}$: has expressed concern about protecting a younger generation}
\item{$t_{81}$: has expressed that a political view is an element of private information for them}
\item{$t_{82}$: has expressed that health status is an element of private information for them}
\item{$t_{83}$: has expressed that identity is an element of private information for them}
\item{$t_{84}$: thinks that posts too much personal/private information}
\item{$t_{85}$: claims that does not post personal/private information}
\item{$t_{86}$: shares personal information in private messages or private account}
\item{$t_{87}$: does not post often}
\item{$t_{88}$: believes that all images are private}
\item{$t_{89}$: believes that some images are more private than others (e.g., they would like to protect only some images or that only some images reveal personal information)}
\item{$t_{90}$: believes that one image can tell a lot (e.g., from one image a lot of information can be inferred)}
\item{$t_{91}$: believes that some photos reveal information that is already shared regardless for other purposes}
\item{$t_{92}$: likes relevant ads / relevant ads are useful}
\item{$t_{93}$: does not like relevant ads}
\item{$t_{94}$: feels that irrelevant ads are useful (e.g., less distraction)}
\item{$t_{95}$: feels that relevant ads are easy to spot (they believe they can easily recognise when something is advertised to them)}
\item{$t_{96}$: feels that relevant ads create an isolation bubble or feel invasive}
\item{$t_{97}$: feels that ads are annoying}
\item{$t_{98}$: believes that monetisation could be ok as long certain conditions are met (e.g., anonymisation, giving consent)}
\item{$t_{99}$: found the number of companies/bids surprising when were exposed to how companies can extract data from their images}
\item{$t_{100}$: does not like that profile data is sold (e.g., makes them feel like a "commodity")}
\item{$t_{101}$: believes that inference could be ok as long certain conditions are met (e.g., anonymisation, giving consent)}
\item{$t_{102}$: has mentioned that it is ok for a company to extract / use the data from the images they post online}
\item{$t_{103}$: has found the inference on images is very shallow (e.g., stereotypical, offensive) when exposed to how companies can extract data from their images}
\item{$t_{104}$: has mentioned that they do not care about inference on what they do share}
\item{$t_{105}$: has mentioned that they do not like inference on their images}
\item{$t_{106}$: believes that it is too late for them to worry when sharing the images online or do not care about what they reveal}
\item{$t_{107}$: has accepted inference performed by the companies on the images that they share online because this is how the world works}
\item{$t_{108}$: initially, believes that not sharing is a safe option when it comes to protecting the personal/private information they share online}
\item{$t_{109}$: initially, believes that image editing could help (e.g., blurring, cropping) when it comes to protecting the personal/private information they share online}
\item{$t_{110}$: initially, believes that profile / post settings manipulations could help (e.g., manually changing the hashtags to the image, setting the account to private, using incorrect location when sharing photos or sharing with delay), when it comes to protecting the personal / private information they share online}
\item{$t_{111}$: after being exposed to how companies can extract data from their images, has expressed that not sharing is a safe option when it comes to protecting the personal / private information they share online}
\item{$t_{112}$: after being exposed to how companies can extract data from their images, has expressed that image editing could help (e.g., blurring, cropping) when it comes to protecting the personal / private information they share online}
\item{$t_{113}$: after being exposed to how companies can extract data from their images, has expressed that profile / post settings manipulations could help (e.g., manually changing the hashtags to the image, setting the account to private, not including the name when sharing the image) when it comes to protecting the personal / private information they share online}
\item{$t_{114}$: after being exposed to how companies can extract data from their images was not sure how to protect the personal / private information they share online}
\item{$t_{115}$: believes that the proposed privacy enhancement filter messes up monetisation based on personal data}
\item{$t_{116}$: believes that the proposed privacy enhancement filter messes up the gathering of personal data and removes objectification}
\item{$t_{117}$: has expressed that the proposed privacy enhancement filter makes the image look better or closer to their taste from their perspective}
\item{$t_{118}$: has expressed that the proposed privacy enhancement filter makes the image look not as good from their perspective}
\item{$t_{119}$: has expressed that the proposed privacy enhancement filter provides general protection and makes them feel better ("adding just to be sure", "they allow to protect me")}
\item{$t_{120}$: believes that the effect of the proposed privacy enhancement filter is very little or none (e.g., "Filters would not change my day-to-day life")}
\item{$t_{121}$: expressed that they would be choosing between the aesthetics of the image and protection when if they were to use a proposed privacy enhancement technology filter}
\item{$t_{122}$: is amazed by the proposed privacy enhancement filter}
\item{$t_{123}$: does not trust the the proposed privacy enhancement filter (agree to use it only when they understand how the privacy enhancement technology works, or believe that the proposed technology is "too good to be true")}
\item{$t_{124}$: believes that privacy is about who has access to image / data inferred from the image}
\item{$t_{125}$: believes that privacy is about what or how much image / data extracted from an image reveals (identities, lifestyle, etc.)}
\item{$t_{126}$: believes that privacy is about a conscious decision of what to share / hide}
\item{$t_{127}$: believes that privacy is about what you might “unintentionally” share (e.g., hidden inference from data)}
\item{$t_{128}$: believes that privacy is about controlling who owns your data}
\item{$t_{129}$: believes that privacy is about to whom image/data is sold}
\item{$t_{130}$: believes that privacy is about controlling what the image is used for (e.g., for example, not allowing it to be a part of the algorithm)}
\item{$t_{131}$: believes that privacy online is deceptive (e.g., information is less private than you think)}
\item{$t_{132}$: believes that privacy online does not exist}
\item{$t_{133}$: believes that privacy is something we do not have full control of}

\end{enumerate}

\onecolumn
\section{Our Privacy Personas' Descriptors}
\label{app:descriptors}
\begin{figure}[h!]
    \centering

    \begin{subfigure}[b]{1\textwidth}
        \centering
        \includegraphics[width = 0.90\linewidth]{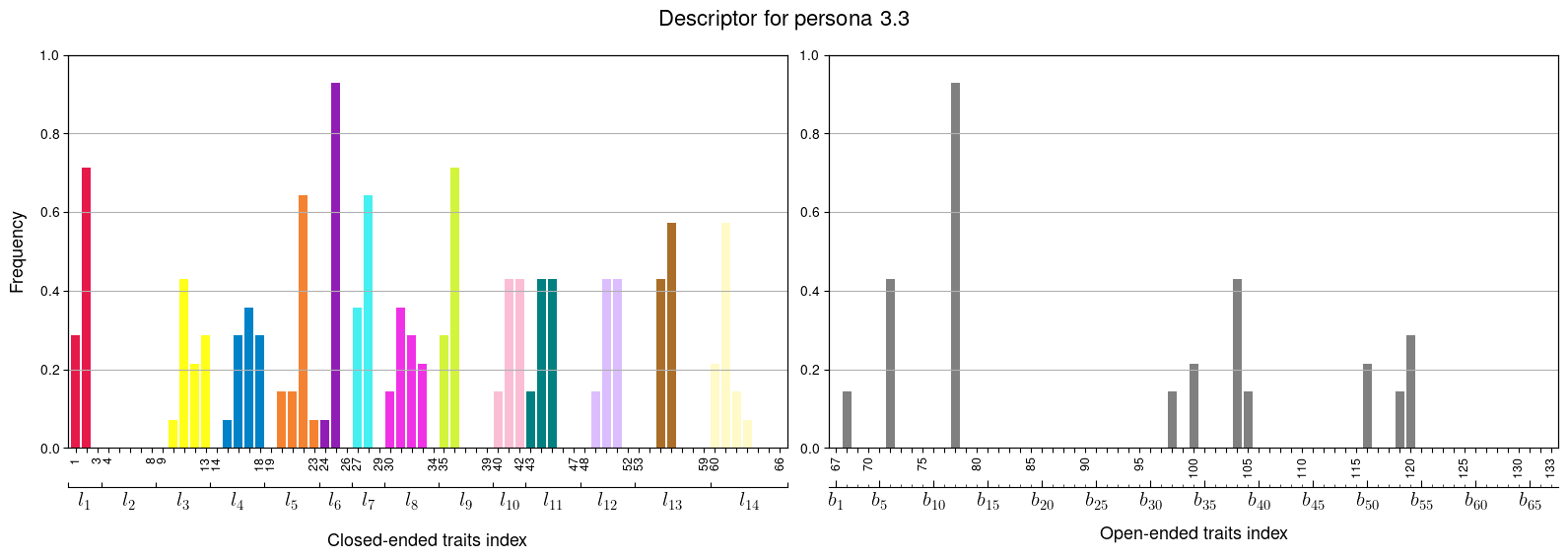}
    \end{subfigure}

    \vspace{20px}
    
    \begin{subfigure}[b]{1\textwidth}
        \centering
        \includegraphics[width = 0.90\linewidth]{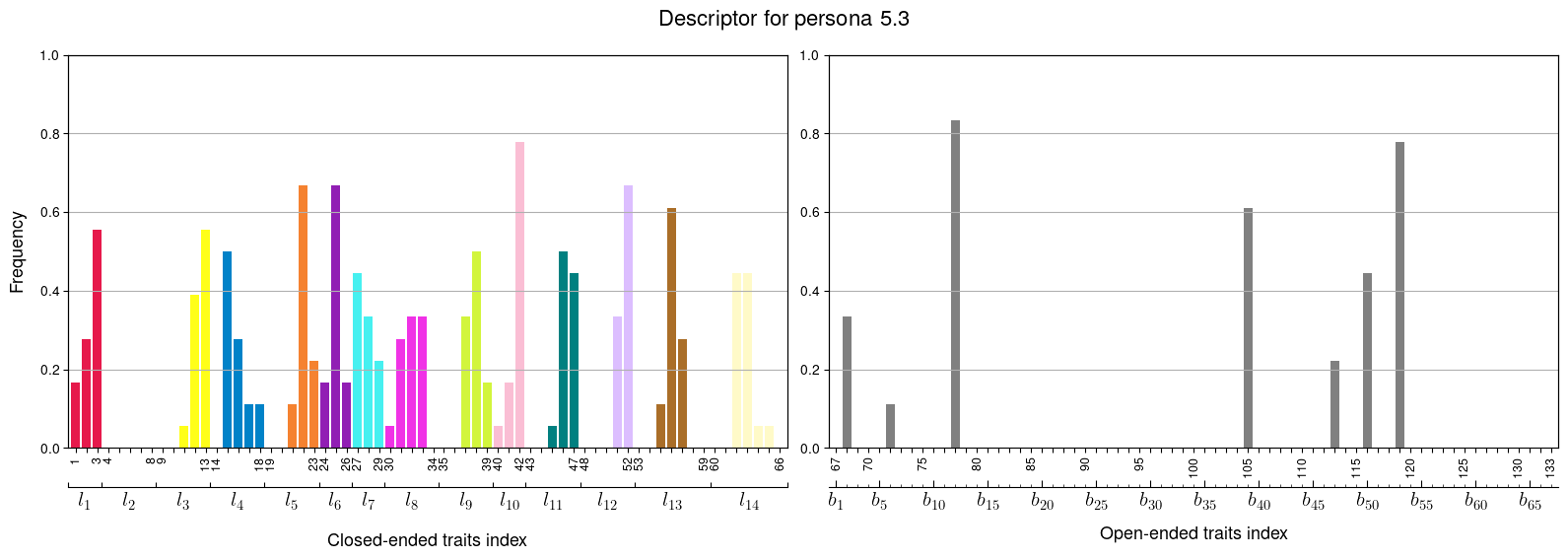}
    \end{subfigure}

    \vspace{20px}

    \begin{subfigure}[b]{1\textwidth}
        \centering
         \includegraphics[width = 0.90\linewidth]{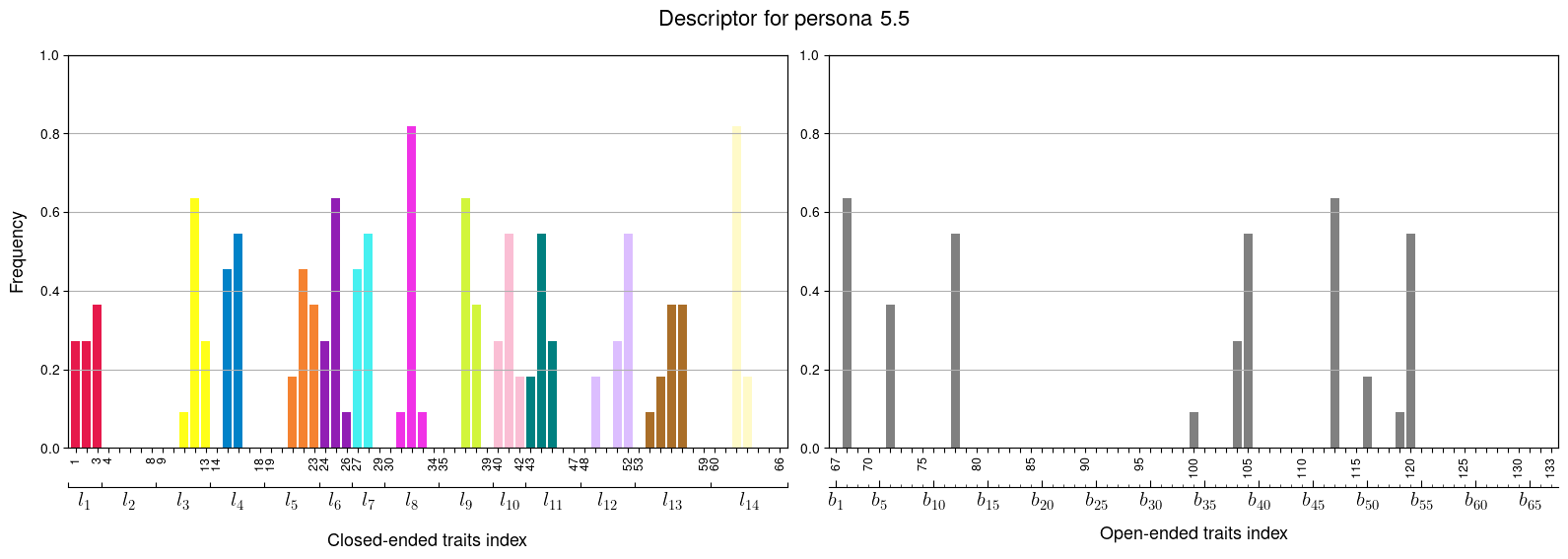}
    \end{subfigure}
    
    \raggedleft{\textit{Fig. \ref{fig:personas_descriptors}  continued on the next page...} \hspace{30px}}
    \caption{ Privacy personas' descriptors for Unconcerned (persona 3.3), In-Control Adopter (persona 5.3) and In-Control Sceptic (persona 5.5). Traits are grouped by the corresponding Likert-scale variables (left) and binary variables (right). The colours of the traits on the left indicate grouping with respect to the Likert-scale variables: $l_1$, ..., $l_{14}$. The upper horizontal axis corresponds to the trait ID, lower horizontal axis corresponds to the explanatory variable ID. }
    \label{fig:personas_descriptors}
\end{figure} 

\begin{figure}[ht]\ContinuedFloat
    \centering

    \begin{subfigure}[b]{1\textwidth}
        \centering
         \includegraphics[width = 0.90\linewidth]{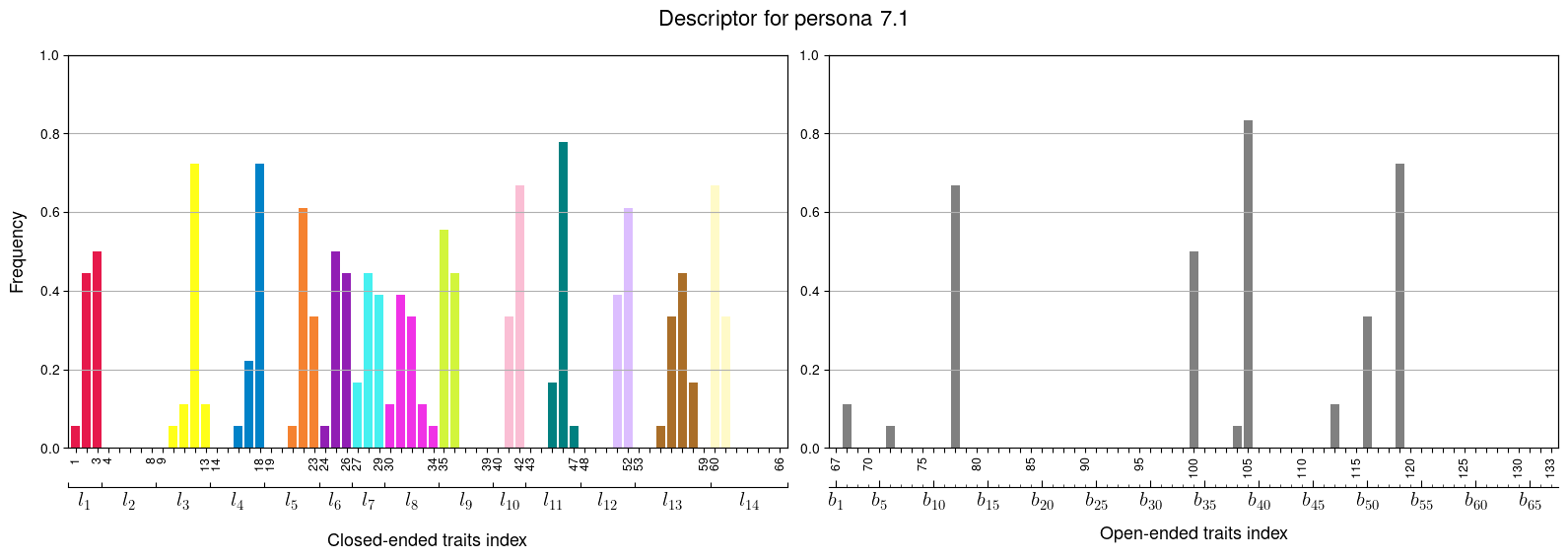}
    \end{subfigure}

    \vspace{20px}

    \begin{subfigure}[b]{1\textwidth}
        \centering
        \includegraphics[width = 0.90\linewidth]{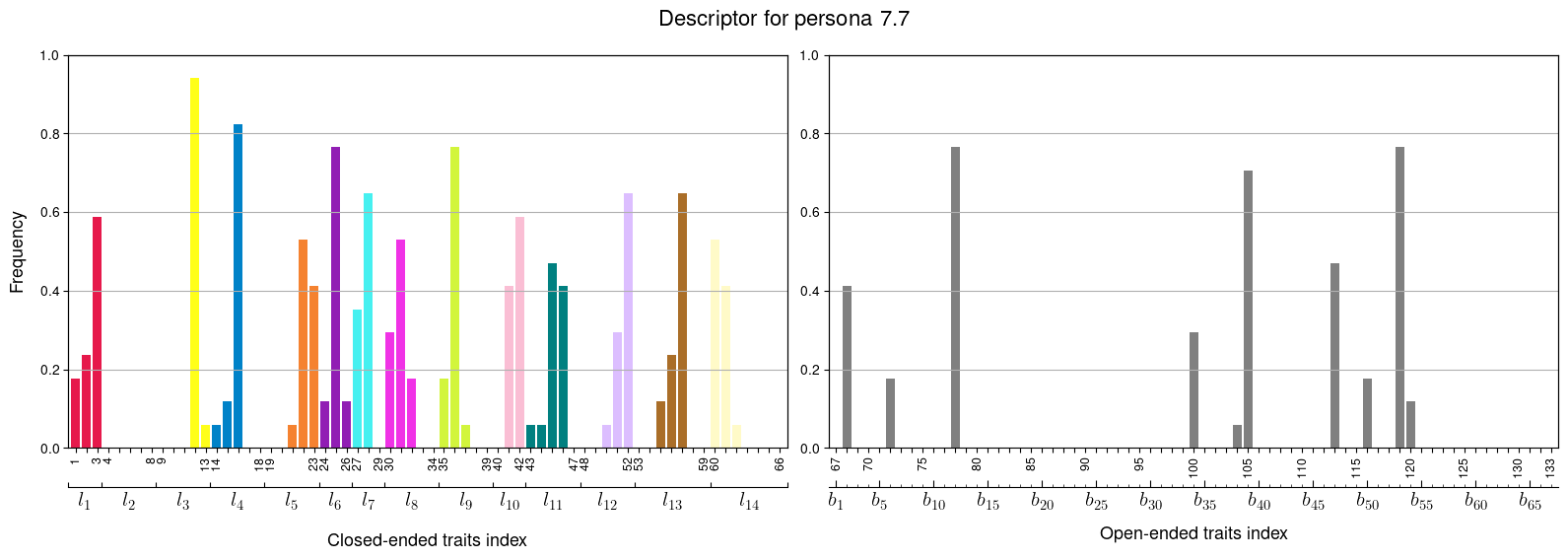}
    \end{subfigure}
    
    \vspace{20px}
    
    \begin{subfigure}[b]{1\textwidth}
        \centering
         \includegraphics[width = 0.90\linewidth]{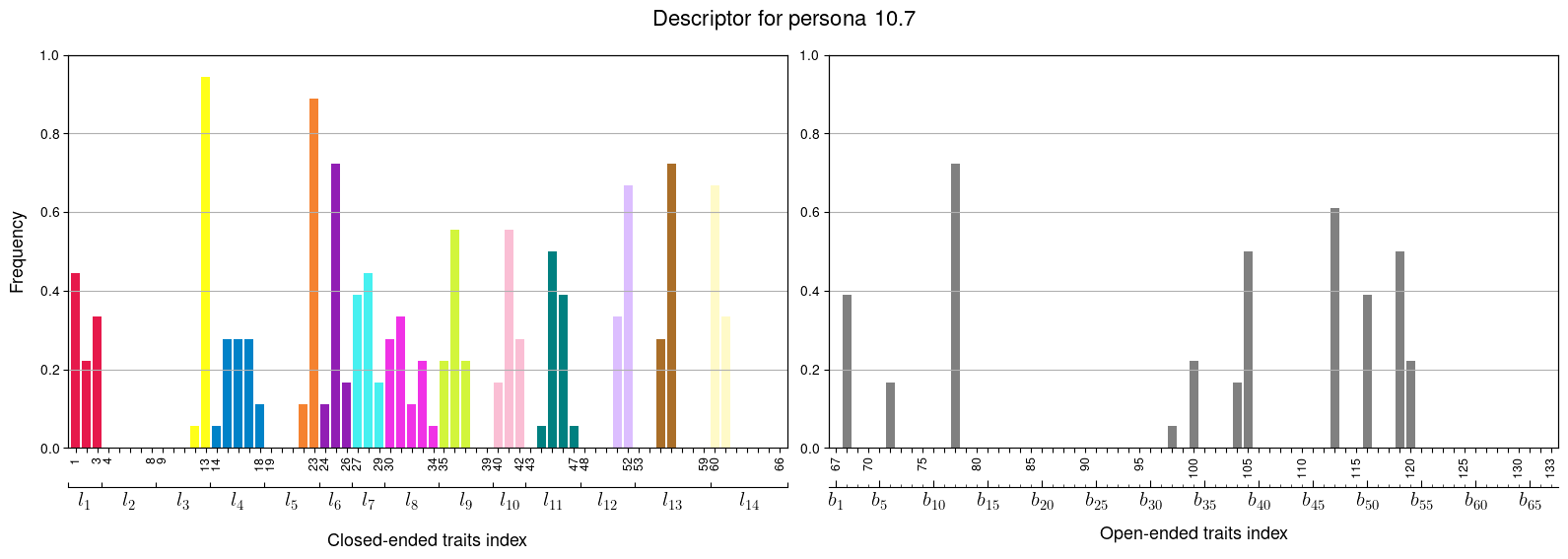}
    \end{subfigure}
    
    \raggedleft{\textit{Fig. \ref{fig:personas_descriptors}  continued on the next page...} \hspace{30px}}
    \caption[]{Privacy personas' descriptors for Knowledgeable Pessimist (persona 7.1), Helpless Protector (persona 7.7) and Occasional Protector (persona 10.7). Traits are grouped by the corresponding Likert-scale variables (left) and binary variables (right). The colours of the traits on the left indicate grouping with respect to the Likert-scale variables: $l_1$, ..., $l_{14}$. The upper horizontal axis corresponds to the trait ID, lower horizontal axis corresponds to the explanatory variable ID. }
\end{figure}

\begin{figure}[ht]\ContinuedFloat
    \centering

    \begin{subfigure}[b]{1\textwidth}
        \centering
        \includegraphics[width = 0.90\linewidth]{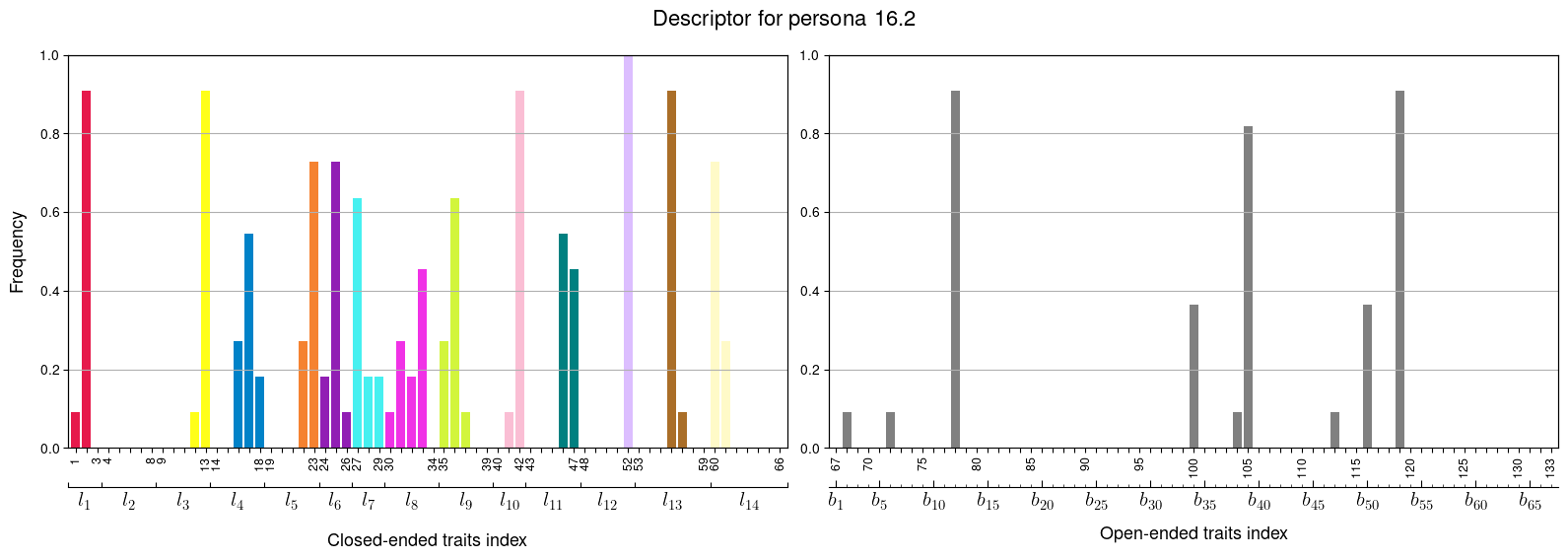}
    \end{subfigure}
    
    \vspace{20px}
    
    \begin{subfigure}[b]{1\textwidth}
        \centering
        \includegraphics[width = 0.90\linewidth]{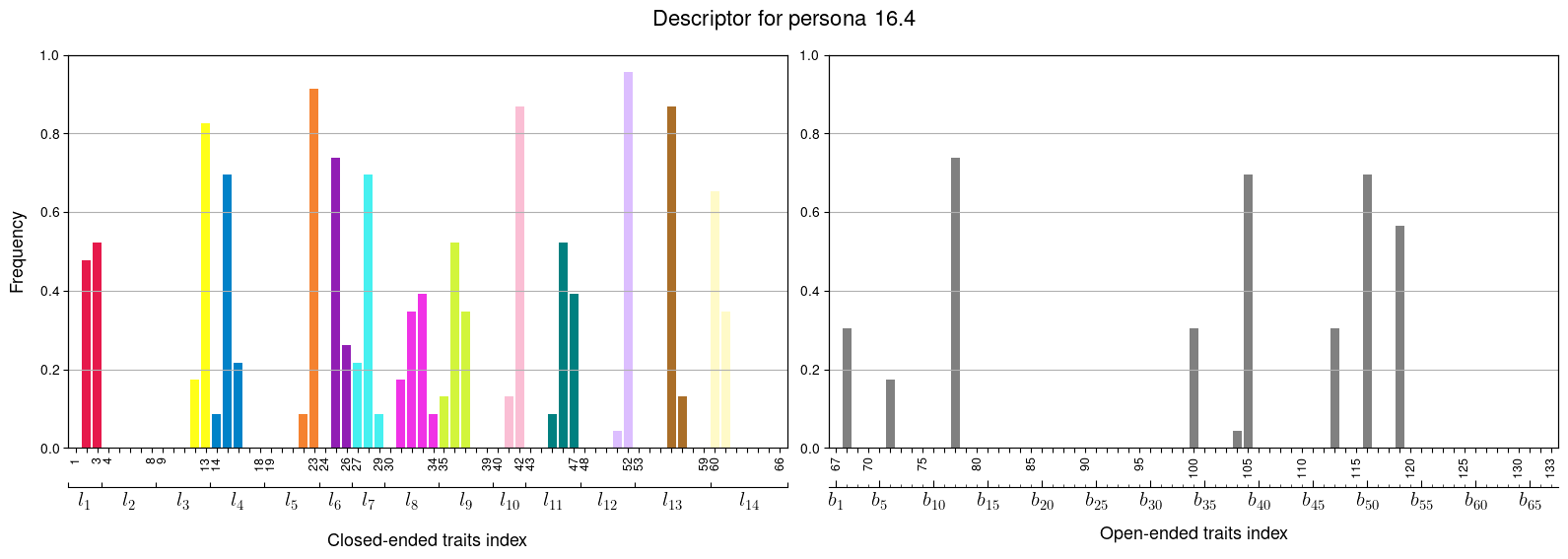}
    \end{subfigure}

    \caption[]{Privacy personas' descriptors for Adopting Protector (persona 16.2) and Knowledgeable Optimist (persona 16.4). Traits are grouped by the corresponding Likert-scale variables (left) and binary variables (right). The colours of the traits on the left indicate grouping with respect to the Likert-scale variables: $l_1$, ..., $l_{14}$. The upper horizontal axis corresponds to the trait ID, lower horizontal axis corresponds to the explanatory variable ID. }
\end{figure} \clearpage

\section{Comparison of Likert-scale Explanatory Variables for Our Privacy Personas}
\label{app:likert_desc}
\begin{figure*}[h!]
    \centering
    \begin{subfigure}[b]{1\textwidth}
        \centering
        \includegraphics[width = 0.26\linewidth]{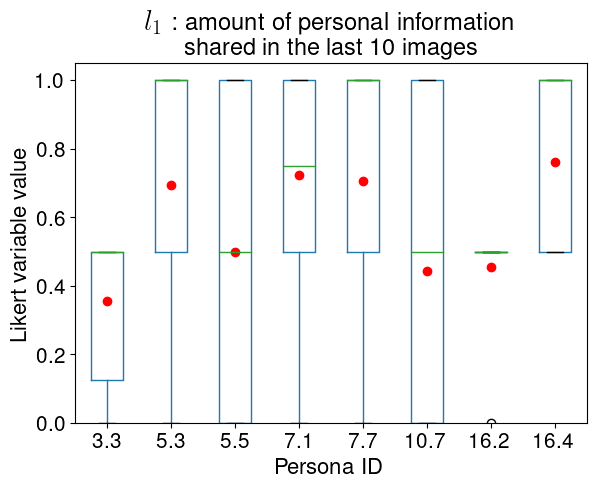}
        \includegraphics[width = 0.26\linewidth]{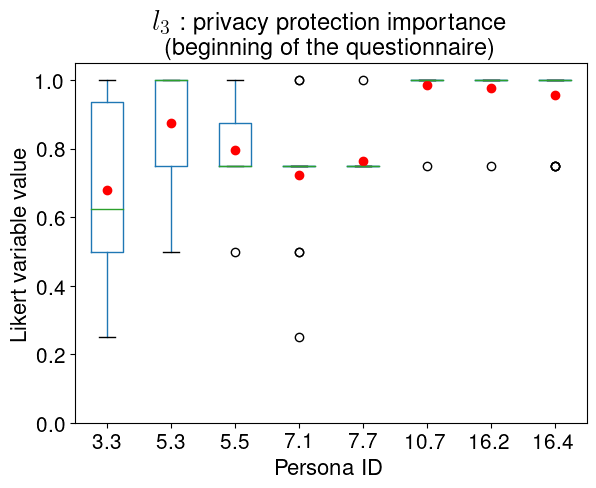}
        \includegraphics[width = 0.26\linewidth]{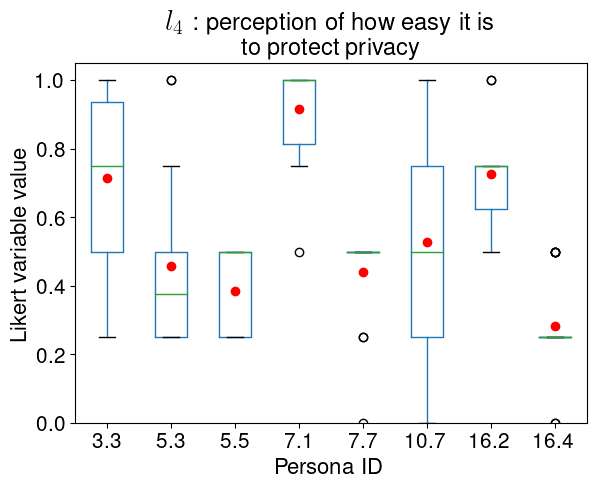}
    \end{subfigure}
    \begin{subfigure}[b]{1\textwidth}
        \centering
        \includegraphics[width = 0.26\linewidth]{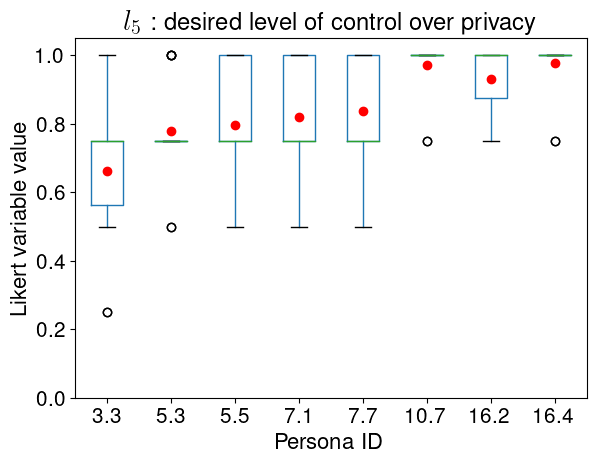}
        \includegraphics[width = 0.26\linewidth]{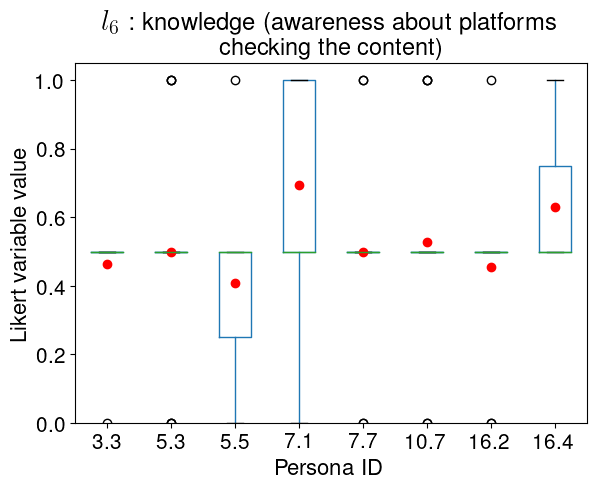}
        \includegraphics[width = 0.26\linewidth]{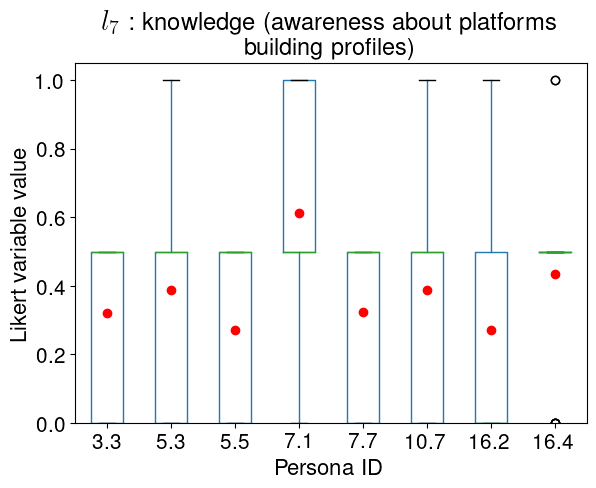}
    \end{subfigure}
    \begin{subfigure}[b]{1\textwidth}
        \centering
        \includegraphics[width = 0.26\linewidth]{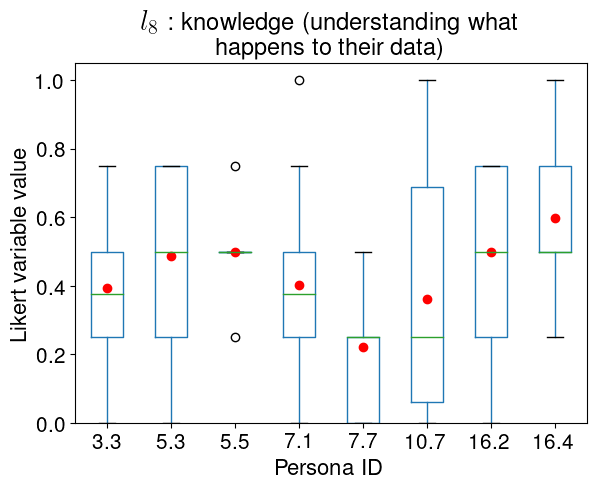}
        \includegraphics[width = 0.26\linewidth]{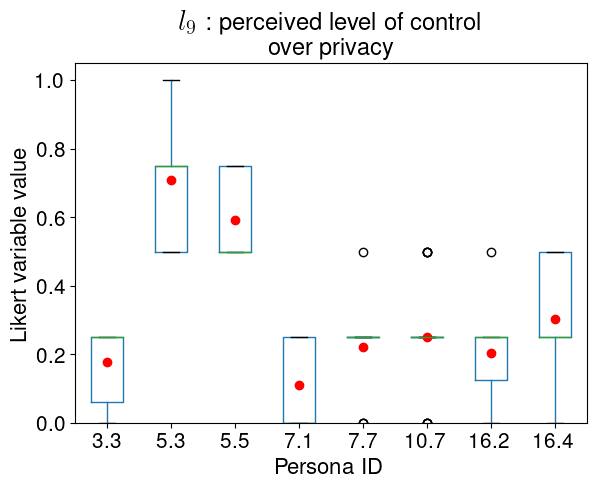}
        \includegraphics[width = 0.26\linewidth]{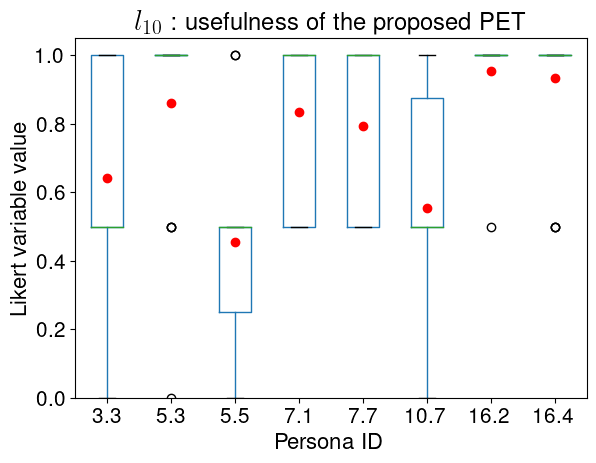}
    \end{subfigure}
    \begin{subfigure}[b]{1\textwidth}
        \centering
        \includegraphics[width = 0.26\linewidth]{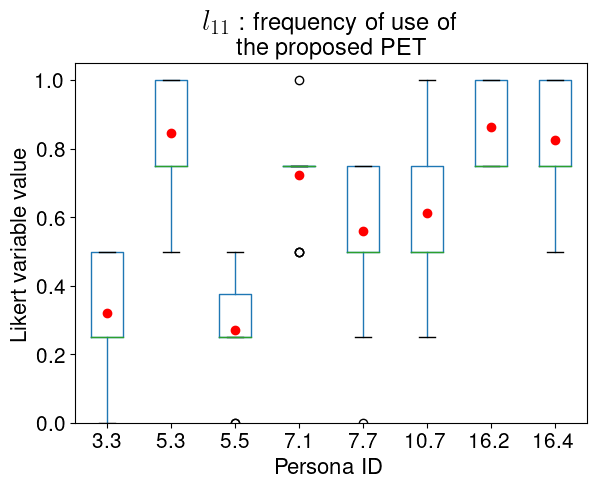}
        \includegraphics[width = 0.26\linewidth]{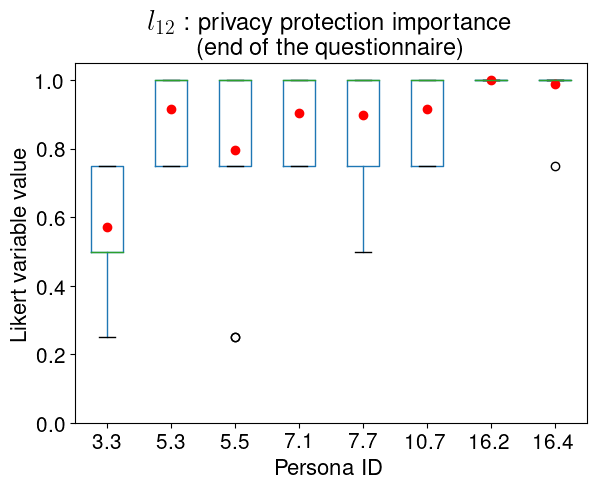}
        \includegraphics[width = 0.26\linewidth]{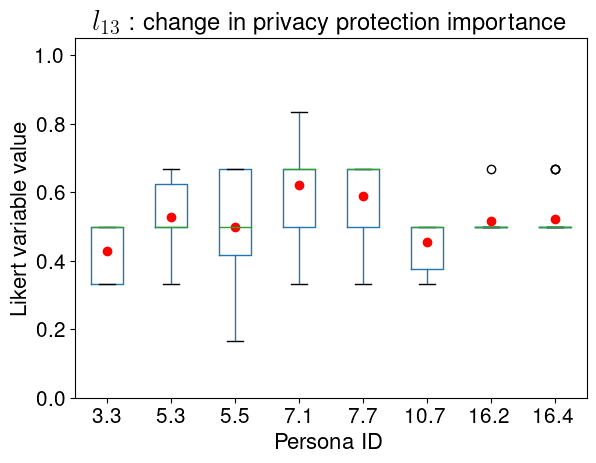}
    \end{subfigure}
    \begin{subfigure}[b]{1\textwidth}
        \centering
        \includegraphics[width = 0.26\linewidth]{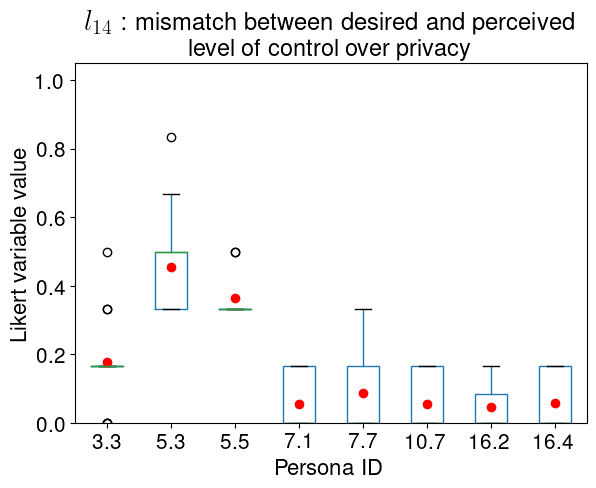}
    \end{subfigure}
    \caption{Distributions of the Likert-scale explanatory variables for personas comparison (the values were re-scaled to $[0, 1]$ range). The red square corresponds to the mean of the explanatory variable. Key – 3.3: Unconcerned, 5.3: In-Control Adopter, 5.5: In-Control Sceptic, 7.1: Knowledgeable Pessimist, 7.7: Helpless Protector, 10.7: Occasional Protector, 16.2: Adopting Protector, 16.4: Knowledgeable Optimist. }
    \label{fig:personas_grouped_by_explanatory_vars_leaves}
\end{figure*}

\clearpage

\section{Our Privacy Personas in 2D Spaces}
\label{app:2D_spaces}
\begin{figure*}[h]
    \centering
    \includegraphics[width = 0.90\linewidth]{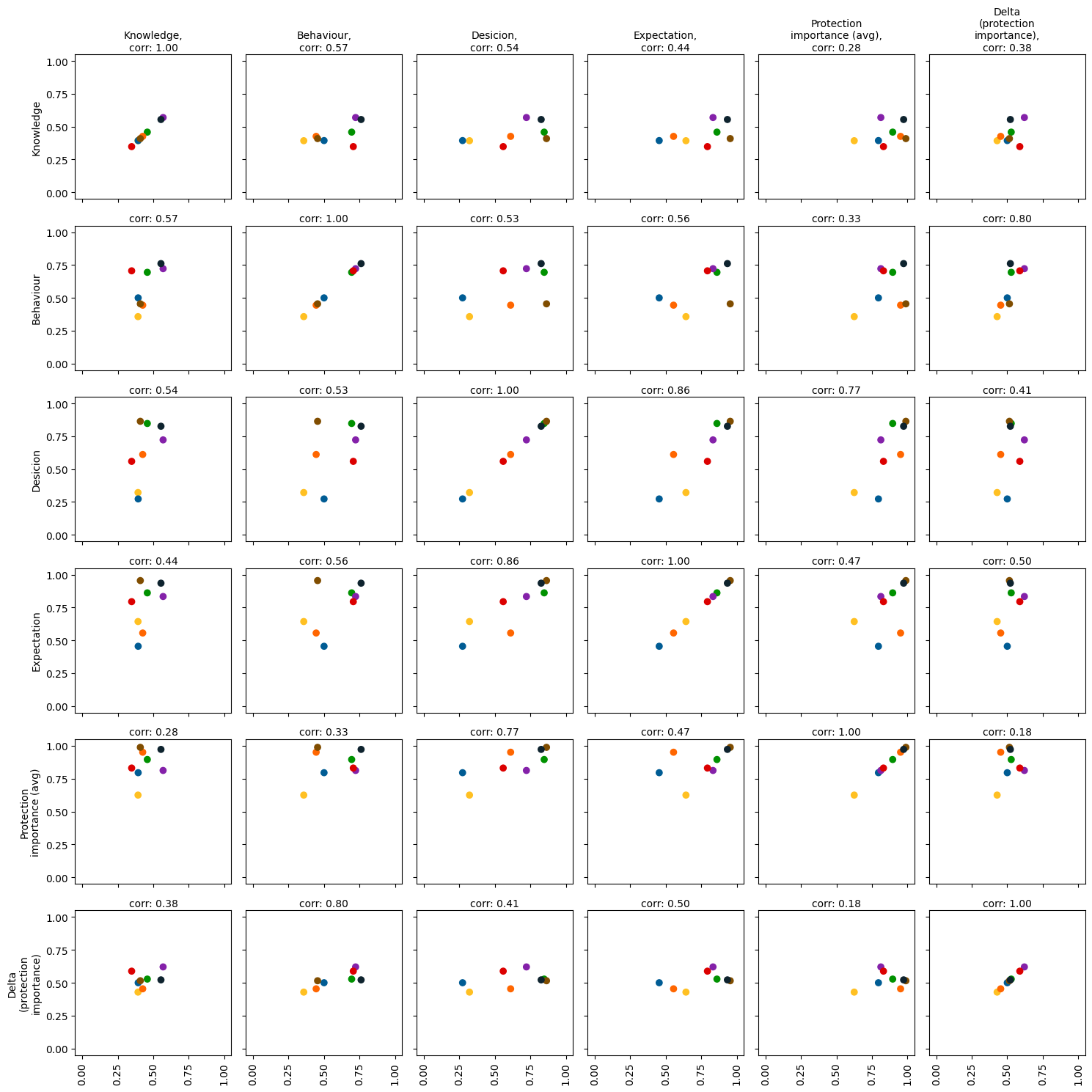}
    \caption{Projection of our privacy personas into 2D spaces, where the dimensions are: knowledge, 
    behaviour, 
    decision to use PET, 
    expectation of PET's efficacy, 
    privacy protection importance, and 
    change in privacy protection importance. 
    Zero/one for knowledge is low/high level of knowledge (maps to $(l_6 + l_7 + l_8)/3$). 
    Zero/one for behaviour is low/high level of self-reported privacy-preserving behaviour (maps to $l_1$). 
    Zero/one for decision to use PET is low/high level of willingness to use PET (maps to $l_{11}$).
    Zero/one for expectation of PET's efficacy is low/high level of perceived efficacy of the proposed PET tool (maps to $l_{10}$).
    Zero/one for privacy protection importance is low/high perceived level of privacy protection importance (maps to $(l_3 + l_{12})/2$).
    Zero/one for change in privacy protection importance is drastic decrease/increase in privacy protection importance throughout the questionnaire (maps to $l_{13}$). 
     Colour-coding: Unconcerned (\tikzcircle[oh_yellow, fill=oh_yellow]{3pt}), 
    In-control Adopter (\tikzcircle[oh_green, fill=oh_green]{3pt}), 
    In-Control Sceptic (\tikzcircle[oh_blue, fill=oh_blue]{3pt}), 
    Knowledgeable Pessimist (\tikzcircle[oh_purple, fill=oh_purple]{3pt}), 
    Helpless Protector (\tikzcircle[oh_red, fill=oh_red]{3pt}), 
    Occasional Protector (\tikzcircle[oh_orange, fill=oh_orange]{3pt}), 
    Adopting Protector (\tikzcircle[oh_brown, fill=oh_brown]{3pt}), 
    Knowledgeable Optimist (\tikzcircle[oh_black, fill=oh_black]{3pt}).
    }
    \label{fig:correlation_likert_personas}
\end{figure*}

\end{document}